\documentclass[10pt,twocolumn,letterpaper]{article}

\usepackage{wacv}              % To produce the CAMERA-READY version

\def\wacvPaperID{42} % *** Enter the WACV Paper ID here
\def\confName{WACV}
\def\confYear{2027}

\usepackage{graphicx}
\usepackage{booktabs}
\usepackage{algorithm}
\usepackage{algorithmic}
\usepackage{multirow}
\usepackage{array}
\usepackage{xcolor}
\usepackage{colortbl}
\usepackage{tcolorbox}

\newcommand{\vlnbert}{VLN$\protect\CircleArrowright$BERT}
\def\CircleArrowright{\ensuremath{%
  \rotatebox[origin=c]{310}{$\circlearrowright$}}}

\usepackage[accsupp]{axessibility}  % Improves PDF readability for those with disabilities.

\usepackage[pagebackref,breaklinks,colorlinks]{hyperref}

\usepackage[capitalize]{cleveref}
\crefname{section}{Sec.}{Secs.}
\Crefname{section}{Section}{Sections}
\Crefname{table}{Table}{Tables}
\crefname{table}{Tab.}{Tabs.}

\usepackage{orcidlink}

\begin{document}

\title{\textsc{NaVIDA}: Vision-Language Navigation with Inverse Dynamics Augmentation}


% Second Author\inst{2,3}\orcidlink{1111-2222-3333-4444} \and
% Third Author\inst{3}\orcidlink{2222--3333-4444-5555}}

\author{Weiye Zhu$^1$\thanks{Equal contribution}\hspace{0.7em}
Zekai Zhang$^{1*}$\hspace{0.7em}
Xiangchen Wang$^{1*}$\hspace{0.7em}
Hewei Pan$^1$\hspace{0.7em} \\
Teng Wang$^1$\hspace{0.7em}
Tiantian Geng$^1$\hspace{0.7em}
Rongtao Xu$^{2}$\hspace{0.7em}
Feng Zheng$^{1,3}$\thanks{Corresponding Author}\\
$^1$ SUSTech
$^2$ MBZUAI
$^3$ SpatialTemporal AI
}

\maketitle

\begin{abstract}

Vision-and-Language Navigation (VLN) requires agents to interpret natural language instructions and act coherently in visually rich environments. Effective navigation requires understanding how actions transform future visual observations. However, most existing methods either learn reactive state-action mappings or model dynamics through future visual prediction. The former lacks explicit action-grounded visual dynamics understanding, while the latter requires predicting high-dimensional visual features that are only indirectly related to navigation decisions.
To address this limitation, we introduce \textsc{NaVIDA} (\textbf{Nav}igation with \textbf{I}nverse \textbf{D}ynamics \textbf{A}ugmentation), an action-centric dynamics learning framework that represents visual transitions directly in the navigation action space with inverse dynamics. By jointly optimizing visual dynamics understanding and instruction-conditioned action prediction in a shared representation, \textsc{NaVIDA} provides structured supervision that improves navigation stability and consistency. To structure this supervision and extend the effective planning range, \textsc{NaVIDA} employs hierarchical probabilistic action chunking (HPAC), which organizes trajectories into multi-step chunks and provides discriminative, longer-range visual-change cues. 
Extensive experiments show that \textsc{NaVIDA} achieves superior navigation performance compared to state-of-the-art methods with fewer parameters (3B vs. 8B). Real-world robot evaluations further validate the practical feasibility and effectiveness of our approach. 
% Code and data will be available upon acceptance. 
Code is available at \href{https://github.com/waynechu1021/NAVIDA}{here}. 
\end{abstract}

\section{Introduction}
\label{sec:intro}
\begin{figure}[t!]
    \centering
    \includegraphics[width=0.75\linewidth]{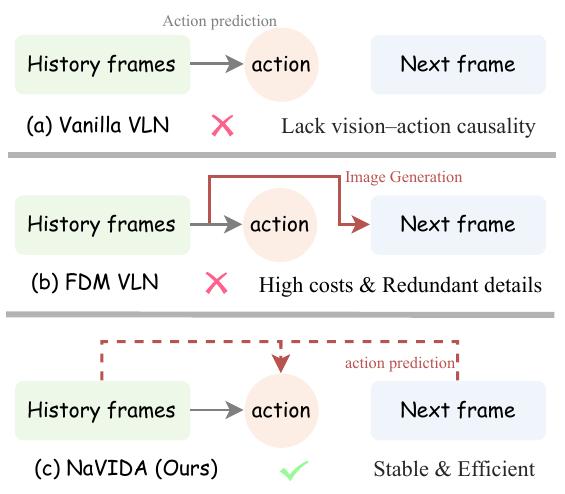}
    \caption{Comparision of three VLN paradigms. (a) Vanilla VLN relies on reactive state-action mapping, lacking the action-grounded visual dynamics; (b) Forward Dynamics Modeling (FDM) adds an additional image generation task which predicts future observations but is compute-heavy and distractible; (c) \textsc{NaVIDA} leverages inverse dynamics supervision to learn action-grounded visual dynamics through action prediction alone, resulting in more efficient and stable training.}
    \label{fig:teaser}
\end{figure}

Vision-and-Language Navigation (VLN)~\cite{r2r,vlnce,vlnpe} is a challenging multimodal sequential decision-making task, requiring an agent to navigate effectively through a 3D visual environment based solely on natural language instructions. 
Accordingly, effective VLN requires not only precise instruction comprehension but also long-horizon temporal modeling and robust generalization to visually diverse, unseen scenes. 
Motivated by advances in video-based Multimodal Large Language Models (MLLMs)~\cite{qwen2_5,nvila,llava-video}, recent studies~\cite{navid,uni-navid,navila,streamvln,monodream,zhang2025activevln,livevln} have integrated these models into VLN to exploit their superior visual perception, semantic understanding, and cross-scene generalization.

% Many existing works~\cite{navid,uni-navid,navila} typically formulate VLN as a short-sighted state-to-action strategy. As shown in ~\cref{fig:teaser}(a), this ``Vanilla VLN'' paradigm directly maps current language instructions and past visual observations to a limited set of subsequent actions. However, this reactive formulation lacks an explicit notion of action-grounded visual dynamics: 
% \textit{how current visual observations are transferred to the next visual observation through the influence of actions}.
% Without internalizing these visual dynamics, the agent merely learns shallow correlations between instructions and observations. Consequently, it cannot anticipate the visual consequences of the executed actions, rendering the navigation process highly susceptible to minor perceptual deviations, which inevitably accumulate over long horizons, leading to severe navigational instabilities, such as stopping too early or uncorrectable trajectory drift.
Many existing methods~\cite{navid,uni-navid,navila} formulate VLN as reactive state-to-action prediction. As shown in~\cref{fig:teaser}(a), this ``Vanilla VLN'' paradigm maps language instructions and visual history directly to subsequent actions, without explicitly modeling how actions transform visual observations. As a result, the agent may rely on shallow instruction-observation correlations and fail to anticipate the visual consequences of its actions, causing small errors to accumulate into premature stopping or trajectory drift.
% To alleviate this issue, recent studies~\cite{monodream,zhang2025cross,huang2025vista} have explored Forward Dynamics Modeling (FDM, as illustrated in~\cref{fig:teaser}b), which forces the agent to predict future visual observations. While FDM explicitly incorporates environmental dynamics into policy learning, it is inherently resource-intensive. Generating high-dimensional future frames often requires complex auxiliary modules (e.g., diffusion-based world models~\cite{chi2025diffusion,cai2025navdp}) and introduces significant computational overhead~\cite{wang2023dreamwalker,zhang2025cross}. More critically, FDM suffers from an inherent objective misalignment: reconstructing dense, pixel-level future frames forces the model to focus on redundant appearance details (e.g., lighting or textures) that are entirely irrelevant to navigation, rather than on optimizing the structured action space.
Recent studies~\cite{monodream,zhang2025cross,huang2025vista} address this issue through Forward Dynamics Modeling (FDM), which predicts future visual observations, as illustrated in~\cref{fig:teaser}(b). 
Note that FDM here is broadly used to denote future-state prediction, rather than strictly the classical action-conditioned formulation.
However, this formulation is computationally demanding because future-observation prediction operates in a high-dimensional visual space and often relies on complex auxiliary components, such as diffusion-based world models~\cite{chi2025diffusion,cai2025navdp, wang2023dreamwalker,zhang2025cross}. More importantly, the future-observation objective is not fully aligned with navigation. Reconstructing dense visual content requires the model to represent appearance details, such as lighting and texture, even when they do not affect the navigation decision. As a result, considerable model capacity may be allocated to visual prediction rather than to action-relevant reasoning.

We argue that action-grounded visual dynamics for navigation does not require reconstructing complete future observations. Based on this insight, we propose \textsc{NaVIDA}, illustrated in~\cref{fig:teaser}(c), an action-centric dynamics learning framework that represents visual transitions directly in the navigation action space through inverse dynamics supervision (IDS). Given visual observations before and after a transition, IDS predicts the action sequence that causes the observed change. This objective encourages the model to capture action-relevant visual differences and align visual transitions with navigation actions. Within \textsc{NaVIDA}, inverse dynamics prediction and instruction-conditioned action prediction are jointly optimized through a shared visual representation and action space. The inverse dynamics objective provides language-agnostic supervision for learning how actions transform visual observations, while the navigation objective grounds these action-aware representations in natural language instructions. This joint formulation promotes representations that support both visual transition understanding and navigation decision-making, reducing reliance on superficial instruction-action correlations. Compared with future-observation prediction, \textsc{NaVIDA} operates in a structured and compact action space, thereby avoiding dense visual reconstruction and the need for a separate future-observation generator.

To better structure this supervision and extend the effective planning horizon, we introduce a Hierarchical Probabilistic Action Chunking (HPAC) mechanism that progressively merges low-level actions into multi-level chunks. This hierarchical and stochastic merging strategy compacts short micro-motions into more informative training targets, providing richer and more discriminative signals for inverse dynamics learning. By introducing controlled randomness into chunk boundaries, HPAC avoids overly rigid compression, preserves segmentation diversity, and improves robustness to minor execution variations. As a result, the agent learns to reason over extended temporal spans while maintaining stable and consistent navigation behavior.
Extensive experiments demonstrate the effectiveness of \textsc{NaVIDA} by achieving state-of-the-art (SOTA) performance on the VLN-CE~\cite{vlnce} benchmark with substantially fewer parameters, establishing a simple and lightweight paradigm for VLN research.

In summary, our contributions are threefold:
\begin{itemize}

% \item We propose \textsc{NaVIDA}, a novel VLN framework that augments policy learning with action-grounded visual dynamics driven by inverse dynamics supervision (IDS). By aligning these two objectives in a shared feature and action space, \textsc{NaVIDA} achieves more stable and consistent navigation.
% \item 

% \item We introduce a Hierarchical Probabilistic Action Chunking (HPAC) mechanism to temporally structure the agent's trajectory into variable-length semantic units. This design improves robustness to minor execution variations, and provides the inverse-dynamics objective with richer visual-transition cues, effectively extending the planning horizon. 

\item We formulate action-grounded visual dynamics learning for VLN as an action-space modeling problem. Instead of predicting future pixels or visual features, we represent visual transitions directly in the navigation action space, providing a more aligned and data-efficient supervision signal for policy learning.

\item We propose \textsc{NaVIDA}, a unified framework that jointly learns action-grounded visual dynamics and instruction-conditioned navigation through inverse dynamics supervision. To provide longer-range and more discriminative transition cues, we further introduce Hierarchical Probabilistic Action Chunking (HPAC), which organizes trajectories into variable-length action chunks, and improves robustness to minor execution variations.

\item Comprehensive experiments demonstrate that \textsc{NaVIDA} achieves SOTA performance with fewer parameters. Real-world tests confirm feasibility, robustness, and faster, more efficient execution, demonstrating the practical benefits of our joint action-understanding approach in physical environments.
\end{itemize}
\section{Related Work}
\label{sec:related work}

\noindent\textbf{Vision-and-Language Navigation.} VLN tasks require an embodied agent to perceive, reason, and act in response to natural-language instructions.
Early studies~\cite{hong2021vln,chen2021history,liu2023bird} focused on discrete navigation on connectivity graphs with panoramic RGB-D inputs, achieving strong in-domain results but relying on heavy sensors and pre-scanned viewpoints.
Subsequent works~\cite{wu2020towards,vlnce} have moved toward continuous control and low-level action prediction to better reflect real-world deployment, while several efforts~\cite{hong2022bridging,wang2023gridmm,an2022bevbert,nguyen2019vision,wang2022towards,schumann2024velma} have investigated monocular variants that relax reliance on panoramic sensing. 
Concurrently, the emergence of MLLMs has spurred monocular-only VLN systems that leverage vision-language backbones for stronger cross-scene transfer and lower hardware cost~\cite{navid,uni-navid,navila,streamvln,pan2026vgp}. Despite these advances, most existing VLN methods remain limited to reactive state-to-action mappings, neglecting explicit modeling of how executed actions relate to subsequent visual transitions. Without such action understanding, agents struggle to anticipate short-horizon observation dynamics, often resulting in unstable turning, premature stopping, and compounding errors over long trajectories.
In contrast, \textsc{NaVIDA} introduces inverse-dynamics supervision as an state-transition objective within policy learning. By aligning this with instruction-based action prediction in a shared feature and action space, our approach improves execution stability under monocular sensing.

% Reactive state-action modeling directly maps historical visual observations and natural language instructions to the action space, lacking causal modeling between changes in visual information and actions. To mitigate this, early works~\cite{mpmvln,vlnpredictfutureview,idmpretraininvln} explored using additional pre-training tasks such as path-mask prediction, image prediction, or image completion. However, these were mostly limited to discrete environmental settings and often relied on extended contextual cues or language instructions. More recent methods~\cite{monodream,zhang2025cross,huang2025vista} attempted to combine forward dynamics modeling and decompose the decision task, predicting future images while simultaneously predicting actions. 
% Although effective in some settings, forward models typically require additional modules for high-dimensional prediction, increase compute and memory, and can bias learning toward redundant appearance details that are weakly aligned with the decision space.
% \textsc{NaVIDA} instead adopts inverse dynamics in a continuous environment: from two adjacent images it directly infers the action that caused the transition, sharing the output space with the reactive policy and avoiding extra generators or decoders. This design reduces optimization difficulty in mapping from image space to action space, improves data efficiency, and injects a causal prior that strengthens instruction grounding, yielding stable long-horizon navigation.
\noindent\textbf{Dynamic Models in Navigation.}
Reactive state-action modeling directly maps historical visual observations and natural language instructions to the action space, without explicitly modeling the dependency between visual transitions and executed actions. To mitigate this, early works~\cite{mpmvln,vlnpredictfutureview,idmpretraininvln} explored using additional pre-training tasks such as path-mask prediction, image prediction, or image completion. However, these were mostly limited to discrete environmental settings and often relied on extended contextual cues or language instructions. More recent methods~\cite{monodream,zhang2025cross,huang2025vista} attempted to combine forward dynamics modeling and decompose the decision task, predicting future images while simultaneously predicting actions to enables modeling of state transitions.
Although effective in some settings, forward models typically require additional modules for high-dimensional prediction, increase compute and memory, and can bias learning toward redundant appearance details that are weakly aligned with the decision space.
\textsc{NaVIDA} instead adopts inverse dynamics in a continuous environment: from two adjacent images it directly infers the action that caused the transition, sharing the output space with the reactive policy and avoiding extra generators or decoders. This design reduces optimization difficulty in mapping from image space to action space, improves data efficiency, and injects an action-grounded transition prior constraints that strengthens instruction grounding, yielding stable long-horizon navigation.

\begin{figure*}[t!]
    \centering
    \includegraphics[width=0.8\linewidth]{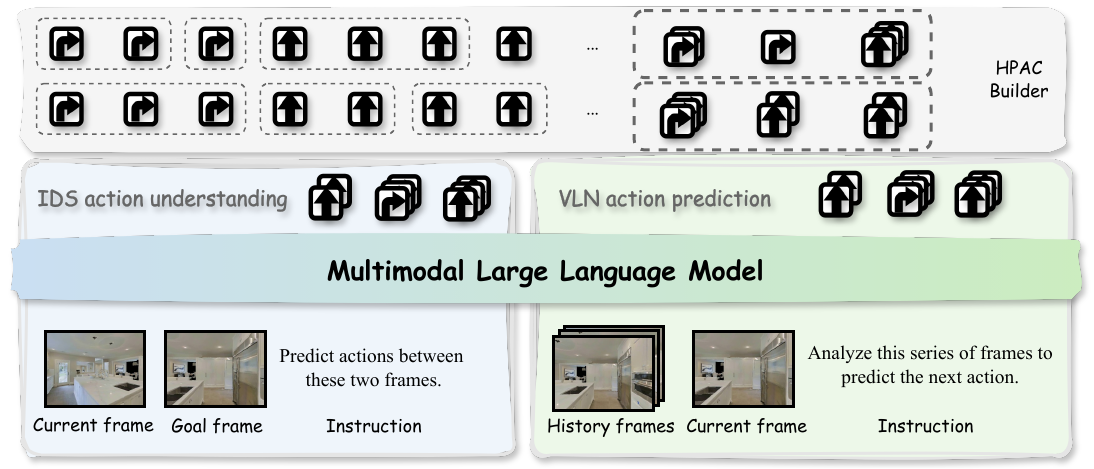}
    \caption{Overview of \textsc{NaVIDA}. A multimodal language-model backbone generates HPAC action blocks. The HPAC mechanism stochastically merges adjacent actions into variable-length motion units, preserving chunk diversity and improves robustness to minor execution variations. Navigation action prediction and inverse dynamics supervision (IDS) are jointly optimized, enabling action-grounded visual dynamics modeling and strengthens action understanding within a shared representation and action space.}
    \label{fig:arch}
\end{figure*}

\section{\textsc{NaVIDA} Framework}
\label{sec:NaVIDA}

\subsection{Preliminary}
\noindent \textbf{Task Formulation.}
\label{sec:task}
At the beginning of each navigation episode, the agent receives a natural language instruction $\mathcal{I}$. Then, at each time step $t$, the agent receives an egocentric RGB observation $v_t$; we denote the observation history by $\mathcal{V}_{0:t}=\{v_0,\ldots,v_t\}$. 
The discrete action space of the navigation agent is:
% \[
\( \mathcal{A}=\{\textsc{Forward},\ \textsc{TurnLeft},\ \textsc{TurnRight},\ \textsc{Stop}\}, \)
% \]
and selecting an action $a_t\in\mathcal{A}$ triggers a transition to a new state and yields the next observation $v_{t+1}$.
Episodes terminate when the agent issues \textsc{Stop} or reaches a maximum step budget $T_{\max}$.
When the agent takes a stop action within 3 meters of the target, the episode is considered successful.

\noindent \textbf{Framework Overview.} 
% Motivated by the need to explicitly couple actions with the visual transitions they induce, we introduce \textsc{NaVIDA} on a standard VLM backbone consisting of a visual encoder, a vision-language projector, and an LLM head (~\cref{fig:arch}).
% On this backbone, our \textsc{NaVIDA} is organized around three mutually consistent components:
% (i) a variable-length action representation via Hierarchical Probabilistic Action Chunking (HPAC),
% (ii) \emph{inverse-dynamics supervision} (IDS) injected at the data level during training, and
% (iii) an \emph{entropy-guided} rule at inference to adaptively choose the executed prefix.
% The chunk grammar aligns IDS and VLN in the same action space; IDS strengthens vision-action causality without adding modules; entropy-guided execution stabilizes decoding by aligning horizon with model confidence.
\textsc{NaVIDA} builds upon a standard Vision–Language Model (VLM) backbone (\cref{fig:arch}). \textsc{NaVIDA} jointly models instruction-based action prediction and action-grounded visual dynamics within a shared feature and action space.
To provide structured and discriminative supervision signals, \textsc{NaVIDA} first introduces Hierarchical Probabilistic Action Chunking (HPAC), which organizes temporally adjacent actions into multi-level chunks and captures longer-range visual transition patterns. Built upon this hierarchical structuring, we incorporate inverse-dynamics supervision (IDS) as an auxiliary objective that trains the model to infer actions from visual transitions, strengthening action understanding and visual–action alignment beyond simple instruction–action correlations.
Through the joint optimization of action prediction and IDS, \textsc{NaVIDA} achieves more stable and consistent navigation under monocular sensing.

\begin{algorithm}[h]
\caption{Hierarchical Probabilistic Action Chunking}
\label{alg:chunking}
\begin{algorithmic}[1]
\REQUIRE Atomic actions $\mathcal{A} = [a_1, \dots, a_N]$, merge prob. $p$, max level-2 chunk size $n$
\ENSURE Hierarchically merged sequence $\mathcal{A}^{(2)}$

\COMMENT{\textbf{Level-1: Probabilistic merging}}
\STATE $\mathcal{A}^{(1)} \leftarrow [\hspace{4pt}]$, $c \leftarrow [a_1]$
\FOR{$i = 2$ to $N$}
    \IF{$a_i = a_{i-1}$ and $\text{len}(c) < 3$ and $\mathcal{U}(0,1) \le p$}
        \STATE Append $a_i$ to $c$
    \ELSE
        \STATE Append $c$ to $\mathcal{A}^{(1)}$; $c \leftarrow [a_i]$
    \ENDIF
\ENDFOR
\STATE Append $c$ to $\mathcal{A}^{(1)}$

\COMMENT{\textbf{Level-2: Deterministic merging}}
\STATE $\mathcal{A}^{(2)} \leftarrow [\hspace{4pt}]$, $c' \leftarrow [\mathcal{A}^{(1)}_1]$
\FOR{$i = 2$ to $|\mathcal{A}^{(1)}|$}
    \IF{$\text{len}(c') < n$}
        \STATE Append $\mathcal{A}^{(1)}_i$ to $c'$
    \ELSE
        \STATE Append $c'$ to $\mathcal{A}^{(2)}$; $c' \leftarrow [\mathcal{A}^{(1)}_i]$
    \ENDIF
\ENDFOR
\STATE Append $c'$ to $\mathcal{A}^{(2)}$

\RETURN $\mathcal{A}^{(2)}$
\end{algorithmic}
\end{algorithm}

\subsection{Hierarchical Probabilistic Action Chunking}
\label{sec:action chunking}
% Instead of predicting a fixed small number of next actions, \textsc{NaVIDA} represents control as variable-length chunks to (a) amplify action-induced visual change for inverse-dynamic supervision and (b) allow farther-ahead decoding when confident, while keeping outputs consistent with the VLN action space.
% We use an HPAC mechanism to merge atomic actions in two stages. In the first probabilistic merging stage, up to $3$ temporally adjacent and visually consistent repetitions of the same atomic action are merged into a sub-chunk with probability $p$, which compacts micro-motions and reduces label noise. In the second, deterministic merging stage, up to $n$ consecutive sub-chunks are merged into a final chunk, yielding a sequence that covers between $n$ and $3n$ atomic actions. The resulting grammar preserves fine-grained short-range decisions while permitting farther-ahead decoding when confidence is high, and it amplifies action-induced appearance and viewpoint changes, thereby producing more discriminative supervision signals. The complete procedure is summarized in Algorithm~\ref{alg:chunking}. This chunk representation is used as the training target, providing a unified label space for IDS and navigation action prediction, and supporting chunk-level uncertainty estimation used by the entropy-guided execution rule.

Instead of predicting a fixed small number of next actions, \textsc{NaVIDA} represents control as variable-length chunks to (a) amplify action-induced visual change for inverse-dynamic supervision, (b) allow farther-ahead decoding when confident, while keeping outputs consistent with the VLN action space and (c) introduces randomness into chunk boundaries to preserve diversity of action chunks and improving robustness to minor execution fluctuations.
We use an HPAC mechanism to merge atomic actions in two stages. In the first probabilistic merging stage, up to $3$ temporally adjacent and visually consistent repetitions of the same atomic action are merged into a sub-chunk with probability $p$. This stochastic merging strategy compacts micro-motions and mitigates label noise, while avoiding the rigid and potentially over-aggressive compression induced by greedy schemes. By introducing controlled randomness into chunk boundaries, it preserves variability in action segmentation and improves robustness to minor execution fluctuations. In the second, deterministic merging stage, up to $n$ consecutive sub-chunks are merged into a final chunk, yielding a sequence that covers between $n$ and $3n$ atomic actions. The resulting grammar preserves fine-grained short-range decisions while permitting farther-ahead decoding when confidence is high, and it amplifies action-induced appearance and viewpoint changes, thereby producing more discriminative supervision signals. The complete procedure is summarized in Algorithm~\ref{alg:chunking}. 
% This chunk representation is used as the training target, providing a unified label space for IDS and navigation action prediction.

% \begin{figure}[t]
%     \centering
%     \includegraphics[width=1.0\linewidth]{fig/data.pdf}
%     \caption{IDM data construction pipeline. From Habitat trajectories, HPAC groups atomic actions into variable-length blocks and induces waypoints; start-end frames yield (current, goal) views, and the intervening actions are ground-truth labels.}
%     \label{fig:data-pipeline}
% \end{figure} 

\subsection{Inverse-Dynamics Supervision}

% To reinforce the causal link between action and visual transition without modifying the architecture, we introduce inverse-dynamics supervision at the data level. Given an HPAC-induced pair of frames $(v_t, v_{t+\ell})$, the model is required to predict the token sequence of the intervening variable-length action chunk $\mathbf{y}_t=(y_{t,1},\ldots,y_{t,n_t})$ using the same action grammar as VLN (i.e., action-type and numeric tokens). Supervision at the level of frame pairs and action chunks aggregates several micro-motions, leading to larger parallax and appearance changes and thus a more discriminative signal in settings with local aliasing (e.g., symmetry or transient occlusion). Following StreamVLN~\cite{streamvln} and JanusVLN~\cite{janusvln}, we construct supervision data using Habitat~\cite{habitat} trajectories from R2R~\cite{r2r}, RxR~\cite{rxr}, ScaleVLN~\cite{wang2023scaling} and DAgger~\cite{dagger}. For each trajectory we use HPAC to define waypoints, and adjacent waypoints define non-overlapping subsequences, where the start frame is stored as the current view, the end frame as the goal view, and the intervening action chunk as the label (~\cref{fig:data-pipeline}). This yields approximately $2.4$M IDS triplets and an additional $\sim\!2.8$M chunked VLN samples built under the same grammar.

To enable action-grounded visual dynamics modeling without modifying the architecture, we introduce inverse-dynamics supervision (IDS) at the data level. Given an HPAC-induced pair of frames $(v_t, v_{t+\ell})$, the model is required to predict the token sequence of the intervening variable-length action chunk $\mathbf{y}_t=(y_{t,1},\ldots,y_{t,n_t})$ using the same action grammar as VLN (i.e., action-type and numeric tokens). In this formulation, IDS serves as an explicit action-understanding objective, encouraging the model to interpret visual transitions in terms of structured action representations, while VLN remains an instruction-conditioned action-prediction task. By sharing the same feature and action space, the two objectives complement each other: IDS strengthens grounded action understanding from visual evidence, and VLN leverages this representation for language-guided decision making.
Supervision at the level of frame pairs and action chunks aggregates multiple micro-motions, resulting in larger parallax and appearance variations and thus providing a more discriminative learning signal in visually ambiguous settings (e.g., symmetry or transient occlusion). Following StreamVLN~\cite{streamvln} and JanusVLN~\cite{janusvln}, we construct supervision data using Habitat~\cite{habitat} trajectories from R2R~\cite{r2r}, RxR~\cite{rxr}, ScaleVLN~\cite{wang2023scaling}, and DAgger~\cite{dagger}. For each trajectory, we apply HPAC to define waypoints, and adjacent waypoints determine non-overlapping subsequences, where the start frame is stored as the current view, the end frame as the goal view, and the intervening action chunk as the supervision label. This process yields approximately $2.4$M IDS triplets and an additional $\sim\!2.8$M chunked VLN samples constructed under the same action grammar.

During training, we jointly optimize the IDS and VLN objectives with shared parameters and task-specific prompts. For IDS, the model predicts $\mathbf{y}_t$ from $(v_t, v_{t+\ell})$ under a task-agnostic prompt $\mathcal{I}_{\mathrm{IDS}}$,
\begin{equation}
\mathcal{L}_{\mathrm{IDS}}^{\tau}=-\sum_{t}\sum_{j=1}^{n_t}\log p\!\left(y_{t,j}^{\star}\,\middle|\, y_{t,<j}^{\star},\, \mathcal{I}_{\mathrm{IDS}},\, v_t,\, v_{t+\ell}\right),
\end{equation}
while for VLN the model predicts the next chunk from the visual history and instruction,
\begin{equation}
\mathcal{L}_{\mathrm{VLN}}^{\tau}=-\sum_{t}\sum_{j=1}^{n_t}\log p\!\left(y_{t,j}^{\star}\,\middle|\, y_{t,<j}^{\star},\, \mathcal{I}_{\mathrm{NAV}},\, v_{<t}\right).
\end{equation}

The combined objective $\mathcal{L}=\sum_{\tau\in D}\big(\mathcal{L}_{\mathrm{VLN}}^{\tau}+\mathcal{L}_{\mathrm{IDS}}^{\tau}\big)$ regularizes the shared representation by encouraging consistency between action-grounded visual dynamics and instruction-conditioned action prediction. Actions inferred from observed transitions are aligned with actions decoded under language guidance, preserving the original VLN action space while avoiding the need to predict high-dimensional future visual appearance.

\begin{table*}[ht]
\caption{Comparison with state-of-the-art methods on VLN-CE R2R Val-Unseen split. Observations used include panoramic view (Pano.), odometry (Odo.), depth sensor (Depth) and single RGB camera (S.RGB). All results are from their respective papers. $*$indicates methods using the waypoint predictor. Best results are highlighted in bold, and the second results are underlined.
}
% \vspace{-1mm}
\centering
\footnotesize
\resizebox{1.47\columnwidth}{!}{
\begin{tabular}{lcc|cccc|cccc}
\toprule
\multirow{2}{*}{Method} &\multirow{2}{*}{Venue} & \multirow{2}{*}{\begin{tabular}[c]{@{}c@{}}{VLM}\\{\# Params}\end{tabular}} & \multicolumn{4}{c|}{Observation} & \multicolumn{4}{c}{R2R Val-Unseen}  \\ 
\cmidrule(lr){4-7} \cmidrule(lr){8-11}
  &  &  & Pano. & Odo. & Depth & S.RGB & NE$\downarrow$ & OS$\uparrow$ & SR$\uparrow$ & SPL$\uparrow$  \\ 
\midrule
HPN+DN$^*$~\cite{krantz2021waypoint}&ICCV2021 & - & $\checkmark$ & $\checkmark$ & $\checkmark$ &  & 6.31  & 40.0  & 36.0  & 34.0  \\
CMA$^*$~\cite{hong2022bridging} &CVPR2022    & -  & $\checkmark$ & $\checkmark$ & $\checkmark$ &  & 6.20  & 52.0  & 41.0  & 36.0  \\
\vlnbert$^*$~\cite{hong2022bridging} &CVPR2022 & - & $\checkmark$ & $\checkmark$ & $\checkmark$ &  & 5.74  & 53.0  & 44.0  & 39.0  \\
Sim2Sim$^*$~\cite{krantz2022sim} &ECCV2022   & - & $\checkmark$ & $\checkmark$ & $\checkmark$ &  & 6.07  & 52.0  & 43.0  & 36.0  \\
GridMM$^*$~\cite{wang2023gridmm} &ICCV2023   & - & $\checkmark$ & $\checkmark$ & $\checkmark$ &  & 5.11  & 61.0  & 49.0  & 41.0  \\
ETPNav$^*$~\cite{an2024etpnav} &TPAMI2024    & -  & $\checkmark$ & $\checkmark$ & $\checkmark$ &  & 4.71  & 65.0  & 57.0  & 49.0  \\ 
ScaleVLN$^{*}$~\cite{wang2023scaling}&ICCV2023  & - & $\checkmark$ & $\checkmark$ & $\checkmark$ &  & 4.80  & --    & 55.0  & 51.0  \\
\midrule
InstructNav~\cite{long2024instructnav}&CoRL2024  & - & - & - & - & - & 6.89 & --   & 31.0 & 24.0  \\
LAW~\cite{law} &EMNLP2021     & -  &  & $\checkmark$ & $\checkmark$ & $\checkmark$ & 6.83  & 44.0  & 35.0  & 31.0   \\
CM2~\cite{cm2}&CVPR2022      & -   &  & $\checkmark$ & $\checkmark$ & $\checkmark$ & 7.02  & 41.5  & 34.3  & 27.6   \\
ETPNav + FF~\cite{eptnavff}&CoRL2024  & -      &  & $\checkmark$ & $\checkmark$ & $\checkmark$ & 5.95  & 55.8  & 44.9  & 30.4   \\
Seq2Seq~\cite{vlnce} &ECCV2020  & -  &  &  & $\checkmark$ & $\checkmark$ & 7.77  & 37.0  & 25.0  & 22.0  \\
NavMorph~\cite{navmorph}&ICCV2025  & -    &  &  & $\checkmark$ & $\checkmark$ & 5.75 & 56.9 & 47.9 & 33.2  \\
\midrule

NaVid~\cite{navid}&RSS2024 & 7B   &  &  &  & $\checkmark$ & 5.47  & 49.1  & 37.4  & 35.9  \\

NaVILA~\cite{navila}&RSS2025 &8B  &  &  &  & $\checkmark$ & 5.22  & 62.5  & 54.0  & 49.0  \\

Uni-NaVid~\cite{uni-navid}&RSS2025 &7B  &  &  &  & $\checkmark$ & 5.58  & 53.3  & 47.0  & 42.7  \\

Aux-Think~\cite{aux-think}&NeurIPS2025 &8B  &  &  &  & $\checkmark$ & 6.08 & 60.0 &  54.8 & 46.9  \\

StreamVLN~\cite{streamvln}&ICRA2026 &7B  &  &  &  & $\checkmark$ & 4.98  & 64.2  & 56.9  & 51.9   \\

MonoDream~\cite{monodream}&AAAI2026 &2B &  &  &  & $\checkmark$ & 5.45 & 61.5 & 55.8 & 49.1  \\
InternVLA-N1(S2)~\cite{internnav2025}&- &7B &  &  &  & $\checkmark$ & 4.89 & 60.6 & 55.4 & 52.1  \\
InternVLA-N1(S1+S2)~\cite{internnav2025}&-&8B &  &  & $\checkmark$ & $\checkmark$ & 4.83 & 63.3 & 58.2 & 54.0  \\
NavFoM~\cite{navfom}&ICLR2026 &7B &  &  &  & $\checkmark$ & 5.01 & 64.9 & 56.2 & 51.2  \\
JanusVLN~\cite{janusvln}&ICLR2026 &8B &  &  &  & $\checkmark$ & 4.78 & \underline{65.2} & \underline{60.5} & \textbf{56.8}  \\
LASAR~\cite{lasar}&CVPR2026 &8B  &  &  &  & $\checkmark$ & -  & 63.1  & 57.0  & 53.9   \\
Progress-Think~\cite{progress-think}&CVPR2026 &4B  &  &  &  & $\checkmark$ & \underline{4.68}  & 63.6  & 60.1  & 53.6   \\
\rowcolor[HTML]{EFF5FB} \textsc{NaVIDA} (ours)&- &3B&  &  &  & $\checkmark$  & \textbf{4.32} & \textbf{69.5} & \textbf{61.4} & \underline{54.7}  \\
\bottomrule
\end{tabular}
}
\label{tab:main result on vlnce r2r}
\end{table*}

\section{Experiment}
\label{sec:experiment}

\begin{table*}[ht]
\caption{Comparison with state-of-the-art methods on VLN-CE RxR Val-Unseen split. 
% Observations used include panoramic view (Pano.), odometry (Odo.), depth sensor (Depth) and single RGB camera (S.RGB). 
All results are from their respective papers. 
% Best results are highlighted in bold, and the second results are underlined.
}
% \vspace{-1mm}
\centering
\footnotesize
\resizebox{1.46\columnwidth}{!}{
\begin{tabular}{lcc|cccc|cccc}
\toprule
\multirow{2}{*}{Method} & \multirow{2}{*}{Venue} & \multirow{2}{*}{\begin{tabular}[c]{@{}c@{}}{VLM}\\{\# Params}\end{tabular}}  & \multicolumn{4}{c|}{Observation} &  \multicolumn{4}{c}{RxR Val-Unseen}  \\ 
\cmidrule(lr){4-7} \cmidrule(lr){8-11} 
  &  &  & Pano. & Odo. & Depth & S.RGB & NE$\downarrow$ & SR$\uparrow$ & SPL$\uparrow$ & nDTW$\uparrow$ \\ 
% \midrule
% CMA$^*$~\cite{hong2022bridging}&CVPR2022   & -   & $\checkmark$ & $\checkmark$ & $\checkmark$ &  &  8.76 & 26.5 & 22.1 & 47.0  \\
% \vlnbert$^*$~\cite{hong2022bridging}&CVPR2022 & - & $\checkmark$ & $\checkmark$ & $\checkmark$ &  & 8.98 & 27.0 & 22.6 & 46.7  \\
% ETPNav$^*$~\cite{an2024etpnav}&TPAMI2024   & -  & $\checkmark$ & $\checkmark$ & $\checkmark$ &  & 5.64 & 54.7 & 44.8 & 61.9  \\ 
% \midrule
% LAW~\cite{law}&EMNLP2021   & -  &  & $\checkmark$ & $\checkmark$ & $\checkmark$  & 10.90 & 8.0 & 8.0 & 38.0  \\
% ETPNav + FF~\cite{eptnavff} &CoRL2024  & -   &  & $\checkmark$ & $\checkmark$ & $\checkmark$  & 8.79 & 25.5 & 18.1  \\
% Seq2Seq~\cite{vlnce}&ECCV2020  & -  &  &  & $\checkmark$ & $\checkmark$  & 12.10 & 13.9 & 11.9 & 30.8  \\
% NavMorph~\cite{navmorph}&ICCV2025  & -  &  &  & $\checkmark$ & $\checkmark$ & 8.85 & 30.8 & 22.8 & 44.2  \\
\midrule
NaVILA~\cite{navila}&RSS2025 & 8B  &  &  &  & $\checkmark$ & 6.77 & 49.3 & 44.0 & 58.8 \\

Uni-NaVid~\cite{uni-navid}&RSS2025 & 7B  &  &  &  & $\checkmark$ & 6.24 & 48.7 & 40.9 & -  \\

Aux-Think~\cite{aux-think}&NeurIPS2025 &8B  &  &  &  & $\checkmark$ & 6.24 & 52.2 & 40.2 & -  \\

StreamVLN~\cite{streamvln}&ICRA2026  & 7B &  &  &  & $\checkmark$ & 6.22 & 52.9 & 46.0 & 61.9  \\

MonoDream~\cite{monodream}&AAAI2026 & 2B &  &  &  & $\checkmark$ &  6.38 & 49.4 & 40.9 & -  \\
InternVLA-N1(S2)~\cite{internnav2025}& - & 7B &  &  &  & $\checkmark$ & 6.41 & 49.5 & 41.8 & 62.6 \\
InternVLA-N1(S1+S2)~\cite{internnav2025}& - & 8B &  &  & $\checkmark$ & $\checkmark$ & 5.91 & 53.5 & 46.1 & \underline{65.3}  \\
NavFoM~\cite{navfom}&ICLR2026 & 7B &  &  &  & $\checkmark$ & \underline{5.51} & \textbf{57.4} & \underline{49.4} & 60.2  \\
JanusVLN~\cite{janusvln}&ICLR2026 & 8B &  &  &  & $\checkmark$ & 6.06 & \underline{56.2} & 47.5 & 62.1  \\
LASAR~\cite{lasar}&CVPR2026  & 8B &  &  &  & $\checkmark$ & - & 52.1 & 47.9 & -  \\
\rowcolor[HTML]{EFF5FB} \textsc{NaVIDA} (ours)&- & 3B &  &  &  & $\checkmark$  & \textbf{5.23} & \textbf{57.4} & \textbf{49.6} & \textbf{67.0}  \\
\bottomrule
\end{tabular}
}
\label{tab:main result on vlnce rxr}
\end{table*}

\definecolor{softred}{HTML}{F28C8C}
\definecolor{softgreen}{HTML}{9BCF9A}
\begin{figure*}[h]
    \centering
    \includegraphics[width=0.85\linewidth]{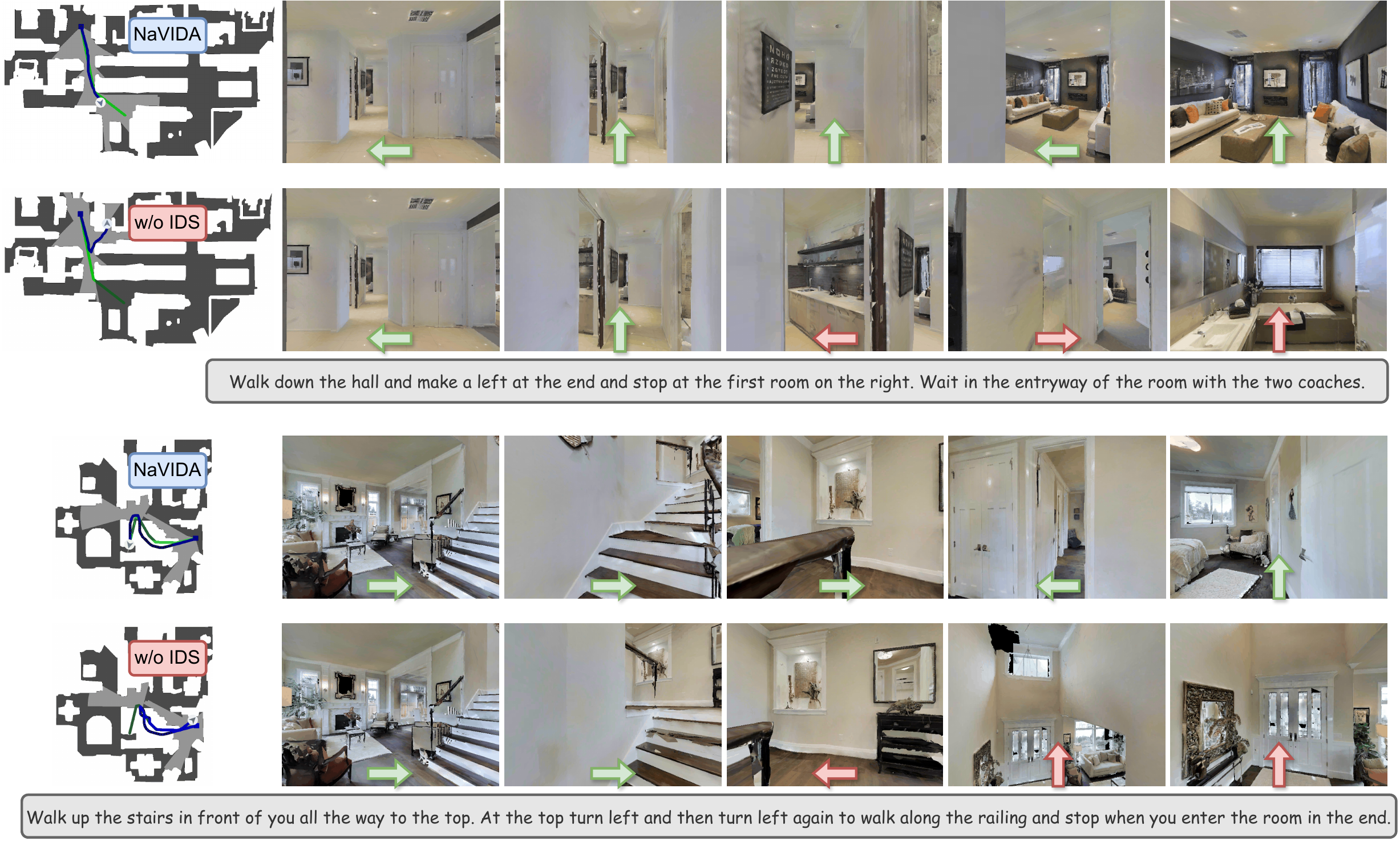}
    \caption{Qualitative results of \textsc{NaVIDA} on VLN-CE R2R Val-Unseen split. Correct actions are marked with \textcolor{softgreen}{green arrows}, while incorrect actions are marked with \textcolor{softred}{red arrows}.}
    \label{fig:simulator_qualitative}
\end{figure*}

\subsection{Experiments Setup}
\noindent \textbf{Implementation Details.}
We implement \textsc{\textsc{NaVIDA}} on top of {Qwen2.5-VL-3B}~\cite{qwen2_5}. During training, the vision encoder is frozen, while the projector and LLM backbone are optimized with a learning rate of $2e-5$ for one epoch over mixed VLN and inverse-dynamics data. Following NaVILA~\cite{navila}, we cap the temporal context at eight historical frames; if a sequence exceeds this limit, we uniformly subsample it to eight frames to preserve coverage across the full trajectory. Unless otherwise stated, HPAC uses merge probability $p=0.7$ and a maximum level-2 chunk size of $n=3$. 
At inference, we only execute the first two predicted action chunks.
Further hyperparameter details are provided in the Supplementary Material.

\noindent \textbf{Simulation Evaluation Benchmarks.}
We evaluate \textsc{NaVIDA} on VLN-CE using the val-unseen splits of R2R-CE~\cite{r2r,vlnce} and RxR-CE~\cite{rxr,vlnce}.
Following standard practice~\cite{janusvln,internnav2025,aux-think}, we report
Navigation Error (NE), Success Rate (SR), Oracle Success rate (OS), Success weighted by Path Length (SPL) and normalized Dynamic Time Warping (nDTW). These metrics jointly capture goal completion, trajectory efficiency, and path fidelity, and allow direct comparison to recent RGB-only and multi-sensor approaches. 
% Detailed explanations of the metrics are provided in the Supplementary Material.

\noindent \textbf{Real-World Evaluation Setup.}
In real-world experiments, we employ two humanoid robot platforms, the AgiBot X2 and the Unitree G1, in an indoor working environment. The AgiBot X2 is equipped with two monocular fisheye RGB cameras; we used only one of them and converted it into an image equivalent to that captured by a pinhole camera. For the Unitree G1, its built-in camera is downward-mounted, and its $55.2^\circ$ field of view (FOV) cannot cover the area directly in front of the robot. Therefore, we additionally mounted a RealSense D455f camera on its chest. When the robot receives a command, it uploads the current RGB image to \textsc{NaVIDA} deployed on a remote server equipped with an NVIDIA 4090 GPU. After receiving the action command from the server, the robot executes the command through the robot's motion API.

\begin{figure*}[h]
    \centering
    \includegraphics[width=0.9\linewidth]{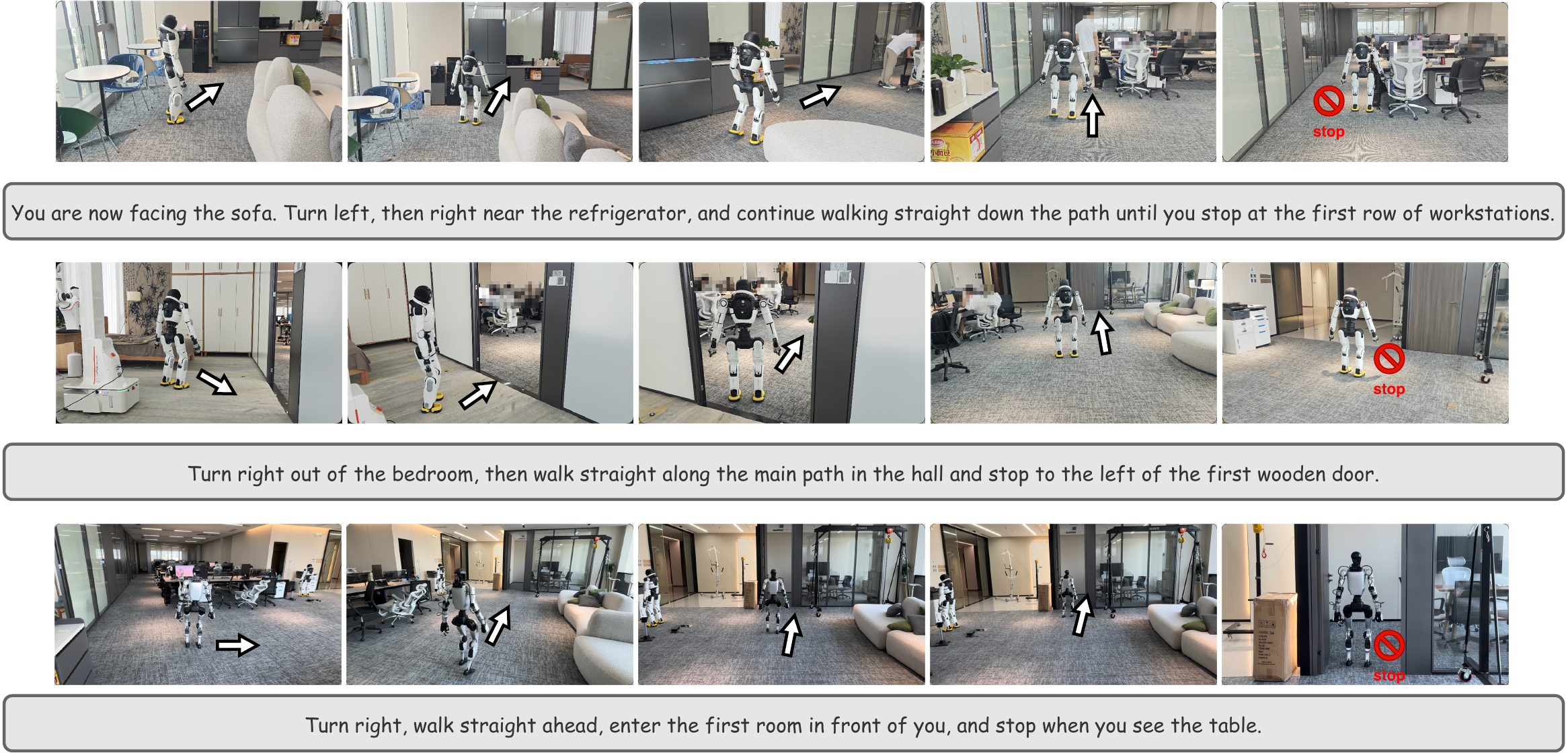}
    \caption{Real world results of \textsc{NaVIDA}. The robot's direction of motion is indicated by arrows in the diagram.}
    \label{fig:realworld_qualitative}
\end{figure*}

\subsection{Main Results}
% \subsection{Comparison on VLN-CE Benchmarks}

\noindent \textbf{Simulation Evaluation.}
As shown in ~\cref{tab:main result on vlnce r2r} (R2R-CE) and ~\cref{tab:main result on vlnce rxr} (RxR-CE), \textsc{NaVIDA} achieves state-of-the-art results on the val-unseen splits using only a single RGB stream and a 3B VLM backbone.
On R2R-CE, \textsc{NaVIDA} attains the best NE/OS/SR and the second-best SPL, surpassing the strongest 8B baseline JanusVLN~\cite{janusvln} by -0.46 NE, +4.3 OS, and +0.9 SR. On RxR-CE, \textsc{NaVIDA} matches the best SR (57.4) and achieves the best NE/SPL/nDTW, improving over StreamVLN and JanusVLN by +1.2 to +4.5 SR, +2.1 to +3.6 SPL, and -0.83 to -0.99 NE. The consistent gains on the longer and linguistically richer RxR-CE benchmark indicate stronger long-horizon alignment and reduced drift, enabled by IDS and hierarchical chunking.
Compared with multi-sensor methods~\cite{navmorph,eptnavff}, \textsc{NaVIDA}’s single-RGB policy remains competitive or superior on core metrics. It also achieves the best performance among RGB-only methods with only ~35\% of the parameters and without auxiliary generators or spatial encoders. Overall, these results demonstrate that inverse-dynamics augmentation with HPAC provides an effective and lightweight alternative to conventional map or forward world modeling.

\cref{fig:simulator_qualitative} contrasts \textsc{NaVIDA} with a variant without IDS where green arrows mark correct actions and red arrows mark errors. 
\textsc{NaVIDA} correctly defers turning until clearing the staircase and selects the intended intersection, whereas the ablated model exhibits premature turns and junction misclassification. These are typical symptoms of error accumulation when the model lacks the understanding of action-grounded visual dynamics and fails to internalize the visual consequences of its actions.

\begin{table}[t]
\centering
\caption{Per-step inference latency on single RTX 4090. ${\dagger}$ denotes results reported by MonoDream.}
\resizebox{\columnwidth}{!}{
\scriptsize
\renewcommand{\arraystretch}{0.9}
\begin{tabular}{lccc}
\toprule
Method & \# Params & Latency / Step (s) $\downarrow$ 
& Relative Latency $\downarrow$ \\
\midrule
MonoDream${\dagger}$~\cite{monodream} & 2B & 0.80 & 2.42$\times$ \\
NaVILA${\dagger}$~\cite{navila}    & 8B & 1.20 & 3.64$\times$ \\
Aux-Think${\dagger}$~\cite{aux-think} & 8B & 1.20 & 3.64$\times$ \\
\rowcolor[HTML]{EFF5FB} \textbf{NAVIDA}      & 3B & \textbf{0.33} & \textbf{1.00$\times$} \\
\bottomrule
\end{tabular}}
\label{tab:latency}
\end{table}

\noindent \textbf{Real-World Evaluation.}
% Due to the lack of standardized real-world evaluation protocols, direct comparison across methods remains difficult. Nevertheless, we conduct 30 navigation trials in a real indoor environment and achieve a success rate of \textbf{19/30} under a 3-meter success threshold, demonstrating the practical applicability of NAVIDA beyond simulation. 
Due to the lack of standardized real-world VLN protocols, direct cross-method comparison remains difficult. We therefore evaluate NaVIDA and StreamVLN under the same indoor setup. Under a 3m success threshold, NaVIDA achieves \textbf{19/30} successful trials. With a stricter 2m threshold, NaVIDA achieves \textbf{16/30}, outperforming StreamVLN’s 12/30, demonstrating improved real-world navigation performance beyond simulation.
\cref{fig:realworld_qualitative} presents examples of real-world trials. \textsc{NaVIDA} executes tight turns at narrow corners and passes smoothly through single-person doorways; even with pedestrians ahead, it resists interference and holds a reliable heading. Taken together, the simulator and real-world evidence identify IDS as the key component behind the observed stability and scene generalization, outperforming the variant without inverse-dynamics supervision.

\noindent \textbf{Efficiency.}
% We compare the inference latency of NAVIDA with recent VLM-based navigation methods on an RTX 4090 (\cref{tab: latency}). Despite achieving competitive navigation performance, NAVIDA requires only \textbf{0.33s} per decision step, making it suitable for real-time embodied navigation.
As shown in Tab.~\ref{tab:latency}, NAVIDA requires only 0.33 s per decision step on an RTX 4090, making it 2.4$\times$ faster than MonoDream~\cite{monodream} and 3.6$\times$ faster than NaVILA~\cite{navila} and Aux-Think~\cite{aux-think}. More importantly, IDS models dynamics directly in the low-dimensional action space, avoiding the high-dimensional panoramic RGB-D feature regression required by MonoDream during training.

\begin{table}[h]
\centering
\resizebox{0.8\columnwidth}{!}{
\scriptsize
\renewcommand{\arraystretch}{0.4}
\begin{tabular}{cc|cccc}
\toprule
HPAC & IDS & NE$\downarrow$ & OS$\uparrow$ & SR$\uparrow$ & SPL$\uparrow$ \\
\midrule
&  & 5.37 & 59.8 & 51.2 & 45.9 \\
$\checkmark$ &  & 5.29 & 61.9 & 52.6 & 47.1 \\
& $\checkmark$ & 5.08 & 62.2 & 54.4 & 48.8 \\
$\checkmark$& $\checkmark$ & 4.91 & 63.2 & 55.5 & 49.6 \\
\bottomrule
\end{tabular}}
\vspace{-2.5mm}
\captionsetup{font=scriptsize}
\caption{Independent ablation of HPAC and IDS.}
\label{tab: independent ablation}
\end{table}

\begin{table}[t]
\centering
\caption{Ablation of IDS Data on VLN-CE R2R Val-Unseen.}
\label{tab:ablation_on_idm}
\setlength{\tabcolsep}{4.5pt}
\renewcommand{\arraystretch}{0.9}
\newcommand{\grayhline}{\noalign{\color{lightgray}\hrule height 0.8pt\color{black}}}
\newcommand{\noidm}{$\circ$}      % 不带 IDM
\newcommand{\withidm}{$\bullet$}  % 带 IDM
% 等宽列类型（垂直/水平都居中）
\newcolumntype{C}[1]{>{\centering\arraybackslash}m{#1}}
\newcommand{\srcw}{1.1cm} % 左侧三列的统一宽度
\resizebox{0.8\linewidth}{!}{
\begin{tabular}{*{3}{C{\srcw}}|cccc}
\toprule
R2R & RxR & DAgger & NE$\downarrow$ & OS$\uparrow$ & SR$\uparrow$ & SPL$\uparrow$ \\
\midrule
\noidm   &         &         & 7.86 & 44.5 & 32.0 & 27.6 \\
\grayhline
\withidm &         &         & 6.99 & 51.5 & 40.5 & 35.5 \\
\grayhline
\withidm & \withidm &        & 5.72 & 57.4 & 47.7 & 41.5 \\
\grayhline
\withidm & \withidm & \noidm & 5.25 & 60.9 & 53.7 & 47.0 \\
\grayhline
\noidm   & \noidm   & \noidm & 5.29 & 61.9 & 52.6 & 47.1 \\
\grayhline
\withidm & \withidm & \withidm & \textbf{4.91} & \textbf{63.2} & \textbf{55.5} & \textbf{49.6} \\
\bottomrule
\end{tabular}}
\hfill
\end{table}

\begin{table}[t]
\centering
\caption{Ablation of the IDS Design Variants on R2R Val-Unseen.}
\label{tab:ablation on ids design}
\setlength{\tabcolsep}{4.5pt}
\renewcommand{\arraystretch}{0.9}
\resizebox{0.6\linewidth}{!}{
\begin{tabular}{c|cccc}
\toprule
Formulation & NE$\downarrow$ & OS$\uparrow$ & SR$\uparrow$ & SPL$\uparrow$ \\
\midrule
w/o IDS & 5.29 & 61.9 & 52.6 & 47.1 \\
MCQ & 5.82 & 56.2 & 49.8 & 45.7 \\
Noised IDS & 6.21 & 53.9 & 46.8 & 42.7 \\
PD & 5.27 & 60.1 & 52.3 & 47.8 \\
Ours & \textbf{4.91} & \textbf{63.2} & \textbf{55.5} & \textbf{49.6} \\
\bottomrule
\end{tabular}}
\end{table}

\subsection{Ablation Study}
\label{sec:ablations}
In this section, we conduct ablations on the effectiveness of IDS, the effectiveness of HPAC and dynamic modeling methods. To ensure efficiency, we conduct experiments on data without ScaleVLN and report the performance on R2R-CE benchmark. Here for convenience, we execute all predicted action chunks in this section. More ablations can be found in Supplementary Material.

\noindent \textbf{Ablation of HPAC and IDS.}
To disentangle the contributions of HPAC and IDS, we independently introduce each component into the baseline. As shown in Tab. \ref{tab: independent ablation}, both HPAC and IDS individually improve performance, while using them together achieves the best results, demonstrating their complementary effects.

\noindent \textbf{Scalability of IDS Data.}
% ~\cref{tab:ablation_on_idm} varies whether R2R, RxR, and DAgger trajectories contribute IDS pairs ($\bullet$) or not ($\circ$). Using IDS consistently improves all metrics: on R2R alone, SR rises from 32.0 to 40.5; with R2R+RxR, SR increases to 47.7; adding DAgger without IDS yields SR 52.6, while enabling IDS across all three reaches SR 55.5 with lower NE (4.91) and higher SPL (49.6). The performance gains are most pronounced in data-scarce scenarios and continue to accumulate as additional data sources are introduced. This pattern suggests IDS injects a transferable, language-agnostic notion of ``how actions change what is seen,'' which complements imitation rollouts such as DAgger.
~\cref{tab:ablation_on_idm} varies whether R2R, RxR, and DAgger trajectories contribute IDS pairs ($\bullet$) or not ($\circ$). 
Incorporating IDS consistently improves performance across all metrics and data configurations. 
% On R2R alone, adding IDS increases SR from 32.0 to 40.5. When extending to R2R+RxR, SR further rises to 47.7. Introducing DAgger without IDS yields an SR of 52.6, while enabling IDS supervision across all three sources achieves 55.5 SR, accompanied by a lower NE (4.91) and higher SPL (49.6).
Notably, the gains are most pronounced in data-scarce settings and continue to accumulate as additional trajectory sources are introduced. This scaling behavior indicates that IDS provides a complementary training signal rather than merely amplifying imitation supervision.
Overall, these results suggest that IDS injects a transferable, language-agnostic notion of action-grounded visual dynamics into the model. Crucially, this capability integrates seamlessly with existing rollout-based paradigms such as DAgger, yielding additive improvements without requiring changes to the underlying imitation framework.

\begin{table}[t]
\centering
\caption{Ablation of HPAC on R2R Val-Unseen.}
\label{tab:ablation_on_chunk_num}
\resizebox{0.9\linewidth}{!}{
\renewcommand{\arraystretch}{0.9}
\begin{tabular}{cc|cccc}
\toprule
Chunk size & Strategy & NE$\downarrow$ & OS$\uparrow$ & SR$\uparrow$ & SPL$\uparrow$ \\
\midrule
1 & Probabilistic & 5.08 & 62.2 & 54.4 & 48.8 \\
2 & Probabilistic & 4.94 & 60.8 & 54.2 & 48.9 \\
3 & Probabilistic & \textbf{4.91} & \textbf{63.2} & \textbf{55.5} & \textbf{49.6} \\
3 & DINOv2-based & 6.29 & 56.1 & 44.7 & 39.5 \\
4 & Probabilistic & 5.23 & 60.2 & 52.5 & 46.4 \\
\bottomrule
\end{tabular}}
\end{table}

\begin{table}[t]
\centering
\caption{Ablation of the dynamic modeling on R2R Val-Unseen.}
\label{tab:ablation_on_dynamic_model}
\newcommand{\grayhline}{\noalign{\color{lightgray}\hrule height 0.8pt\color{black}}}
\renewcommand{\arraystretch}{0.9}
\resizebox{0.8\linewidth}{!}{
\begin{tabular}{c|cccc}
    \toprule
    Dynamic Method  &  NE$\downarrow$ & OS$\uparrow$ & SR$\uparrow$ & SPL$\uparrow$ \\
    \midrule
      Base      & 5.29 & 61.9 & 52.6 & 47.1 \\
    \grayhline
       w/ FDS (epoch-1)     & 6.13  & 56.0 & 44.6 & 38.7 \\
       w/ FDS (epoch-2)     & 5.06  & 60.1 & 51.0 & 45.2 \\
       w/ FDS (epoch-3)     & 5.06  & 62.4 & 53.9 & 48.2 \\
    \grayhline
       w/ IDS      & \textbf{4.91} & \textbf{63.2} & \textbf{55.5} & \textbf{49.6} \\
      \bottomrule
    \end{tabular}}
\end{table}

\noindent \textbf{IDS Design.} 
To better understand how IDS facilitates action prediction, we conduct ablations on its construction strategy (\cref{tab:ablation on ids design}).
First, we corrupt IDS by replacing ground-truth actions with random ones during data construction (Noised IDS), breaking the alignment between actions and visual transitions while keeping the VLN data and training scale unchanged. This results in a substantial performance drop, indicating that IDS benefits navigation not by increasing data size, but by providing consistent supervision on action–visual dynamics.
We further reformulate IDS as a multiple-choice question (MCQ), where the model selects the correct action from four candidates instead of directly generating the action token. Although the action–visual relationship is preserved, performance slightly degrades. We attribute this to the mismatch between training and inference: the MCQ formulation allows the model to rely on candidate comparison, whereas generative IDS requires direct action prediction from visual changes. Sharing the same generative action space with VLN therefore enables the learned transition knowledge to transfer more effectively to navigation.
We also implement a privileged distillation (PD) variant. A teacher model trained with IDS supervision transfers its knowledge to a student through KL-divergence minimization on action distributions. Despite the additional supervision, PD performs similarly to the baseline and remains inferior to \textsc{NaVIDA}. We hypothesize that PD transfers dynamics knowledge only through output-level imitation, treating it as a fixed signal after training. In contrast, \textsc{NaVIDA} jointly optimizes IDS and VLN within a unified framework, allowing action–visual alignment to be continuously reinforced through shared representations and directly influence navigation reasoning, resulting in more effective learning. Beyond the controlled ablations above, we further analyze the relationship between IDS consistency and navigation capability in the Supplementary Material.

\noindent \textbf{Chunk Size of HPAC.}
\cref{tab:ablation_on_chunk_num} presents an ablation study of HPAC on the VLN-CE R2R val-unseen split. When the maximum number of action chunks is set to $1$ with merging probability fixed to $1$, HPAC reduces to the deterministic action-merging strategy used in prior work\cite{navid,navila}. The best performance is achieved with up to $3$ action chunks under our probabilistic merging scheme, while fewer or more chunks degrade results. This suggests that moderate chunk extension benefits longer-horizon planning, whereas overly long chunks may exceed the model’s effective perception range. Importantly, the probabilistic merging strategy introduces controlled randomness into chunk boundaries, preserving diversity of action chunks and improving robustness to minor execution fluctuations. 
We further evaluate a deterministic DINOv2-based~\cite{dinov2} strategy that groups adjacent frames by visual similarity. This variant performs worst overall, likely because DINOv2 captures semantic similarity rather than the geometric and viewpoint changes critical for navigation. For instance, observing the same object from different viewpoints may yield high DINOv2 similarity despite significant spatial displacement. As a result, semantically driven merging can obscure motion cues and weaken action supervision. Developing geometry-aware grouping strategies based on optical flow or multi-frame motion consistency is a promising direction for future work.

\noindent \textbf{Inverse vs. Forward Dynamics.}
% To evaluate the effectiveness of inverse dynamics supervision, we compare it with forward dynamics supervision (FDS), as both aim to enhance action-grounded visual dynamics modeling. Following MonoDream~\cite{monodream}, FDS predicts future visual features using the same VLM without introducing additional modules. Considering its slower convergence, we train FDS for three epochs to assess both effectiveness and efficiency relative to IDS.
% As shown in~\cref{tab:ablation_on_dynamic_model}, IDS consistently outperforms both FDS and the baseline without dynamic supervision, while FDS initially causes noticeable performance degradation. We attribute this gap to differences in output space alignment and optimization difficulty. Predicting dense future visual features requires modeling high-dimensional appearance variations, which is challenging for a pre-trained VLM and typically demands more training (e.g., MonoDream trains for five epochs). In contrast, IDS shares the same output space as action prediction in VLN, making it better aligned with the model’s pre-training and downstream objectives.
% Although FDS gradually improves with additional training and eventually surpasses the baseline after three epochs, it still lags behind IDS, which achieves superior performance after only one epoch. These results indicate that IDS provides more stable and data-efficient supervision, whereas FDS requires longer optimization to learn reliable future-state predictions.
To evaluate the effectiveness of inverse dynamics supervision, we compare IDS with forward dynamics supervision (FDS), as both aim to model action-grounded visual dynamics. Following MonoDream~\cite{monodream}, FDS predicts future visual features using the same VLM without additional modules. Given its slower convergence, we train FDS for up to three epochs to assess both effectiveness and efficiency.
As shown in~\cref{tab:ablation_on_dynamic_model}, IDS consistently outperforms both FDS and the baseline without dynamic supervision. We attribute this advantage to better output-space alignment: IDS shares the same action prediction space as VLN, whereas FDS requires modeling high-dimensional future visual features, a more challenging objective that is less aligned with the model’s pre-training. Although FDS gradually improves with additional training and eventually surpasses the baseline, it still remains inferior to IDS. Notably, IDS achieves superior performance after only one epoch, demonstrating more stable and data-efficient learning.
\section{Conclusion}
\label{sec:conclusion}

In this work, we presented \textsc{NaVIDA}, a unified VLN framework that integrates policy learning with action-grounded visual dynamics modeling. Through chunk-based inverse-dynamics supervision, \textsc{NaVIDA} achieves more interpretable and stable navigation. The proposed hierarchical probabilistic action chunking mechanism structures navigation trajectories into multi-level motion units, providing richer and longer-range supervision for inverse dynamics learning. Extensive experiments demonstrate that \textsc{NaVIDA} achieves superior navigation performance with fewer parameters compared to state-of-the-art methods, and real-world robot evaluations confirm its effectiveness and practical applicability.

\clearpage  % TODO FINAL: This \clearpage needs to be removed from both review and camera-ready versions.

% \section*{Acknowledgements}
% Please insert your acknowledgments here.

\bibliographystyle{splncs04}
\bibliography{main}

% \title{
% \texorpdfstring{
% \textsc{NaVIDA}: Vision-Language Navigation with Inverse Dynamics Augmentation\\
% {\normalfont\normalsize \textbf{Supplementary Material}}
% }{
% NaVIDA: Vision-Language Navigation with Inverse Dynamics Augmentation (Supplementary Material)
% }}

\clearpage
\setcounter{page}{1}
\maketitlesupplementary

% % \maketitle
% \setcounter{page}{1}
% \appendix

% \cftsetindents{section}{0em}{1.5em}
% \cftsetindents{subsection}{1em}{2.5em}
% \cftsetindents{subsubsection}{2em}{3.5em}
% \section*{Table of Contents}
% \startcontents[appendix]
% \printcontents[appendix]{l}{1}{\setcounter{tocdepth}{3}}

\section{Prompts for Different Tasks}
\label{sec:app_promot}

We use task-specific prompts for VLN and for inverse-dynamics supervision, both expressed in the same action grammar (action-type + numeric tokens).

For vision-language navigation task, we use the following prompt template to instruct the model to predict the corresponding action chunks:

\begin{tcolorbox}[
  colback=gray!15,    % 背景颜色 (e.g., gray!10 for 10% gray, lightgray)
  colframe=black!75,  % 边框颜色 (e.g., black, darkgray)
  boxrule=0.5pt,      % 边框线条粗细
  arc=1mm,            % 边框圆角半径 (0mm for sharp corners)
  % title=Prompt Example, % 如果需要标题，取消注释这一行
]
Imagine you are a robot programmed for navigation tasks. You have been given a video of historical observations: \textless image\textgreater,...,\textless image\textgreater and current observation: \textless image\textgreater. Your assigned task is: [Instruction]. Analyze this series of images to decide your next move, which could involve turning left or right by a specific degree, moving forward a certain distance.
\end{tcolorbox}

Among them, [Instruction] is the language instruction given for the current navigation task. For inverse-dynamics supervision, we use prompt a prompt template like the following:
\begin{tcolorbox}[
  colback=gray!15,  
  colframe=black!75,  
  boxrule=0.5pt,      
  arc=1mm,            
  % title=Prompt Example, 
]
Imagine you are a robot programmed for navigation tasks. You have been given an image of current view \textless image\textgreater and an image of the goal view \textless image\textgreater. Analyze the two images to predict the navigation action that would move the robot from the current viewpoint to the goal view, which could involve turning left or right by a specific degree or moving forward a certain distance.
\end{tcolorbox}

For the multiple-choice question (MCQ) of IDS, the prompt is like follows:

\begin{tcolorbox}[
  colback=gray!15,  
  colframe=black!75,  
  boxrule=0.5pt,      
  arc=1mm,            
  % title=Prompt Example, 
]
Imagine you are a robot programmed for navigation tasks. You have been given an image of current view \textless image\textgreater and an image of the goal view \textless image\textgreater. Analyze the two images to predict the navigation action that would move the robot from the current viewpoint to the goal view. Select the correct answer from the four options (A, B, C, or D). Options: \textbackslash n A. \textless action A\textgreater \textbackslash n B. \textless action B\textgreater \textbackslash n C. \textless action C\textgreater \textbackslash n D. \textless action D\textgreater
\end{tcolorbox}

For the forward-dynamics supervision, the requirement is to predict the visual observation features of the next time step based on the given historical frames and language instructions. We use the following prompt:

\begin{tcolorbox}[
  colback=gray!15,  
  colframe=black!75,  
  boxrule=0.5pt,      
  arc=1mm,            
  % title=Prompt Example, 
]
Imagine you are a robot programmed for navigation tasks. You have been given a video of historical observations: \textless image\textgreater,...,\textless image\textgreater and and current observation: \textless image\textgreater. Your assigned task is: [Instruction]. Analyze this series of images to predict the future observation.
\end{tcolorbox}

\section{More Implementation Details}
\label{sec:app_implementation_details}

\subsection{Data Details}
\label{sec:app_data_analysis}

\begin{figure}[h]
    \centering
    \includegraphics[width=1.0\linewidth]{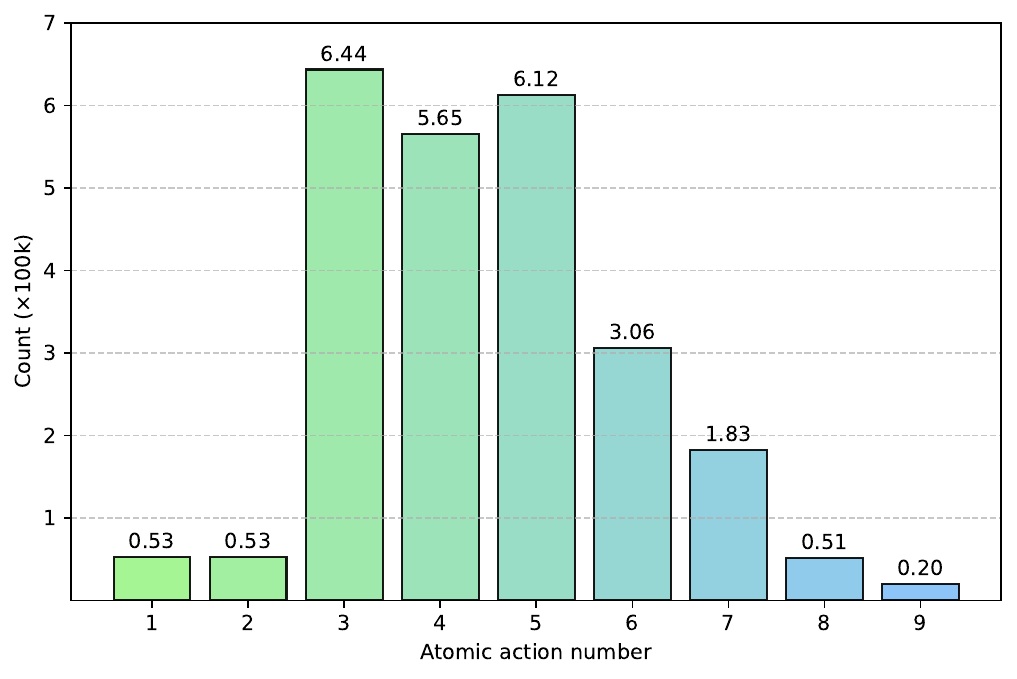}
    \caption{Distribution of atomic actions per HPAC chunk in the IDM data. Most chunks contain 3-5 atomic actions.}
    \label{fig:data_distribution}
\end{figure}

\cref{fig:data_distribution} depicts the distribution of atomic actions per HPAC chunk in our inverse-dynamics (IDM) dataset. Although most chunks contain $3$ to $5$ atomic actions, which consistent with the fixed chunk sizes used in prior works~\cite{navid,navila,streamvln}, a non-negligible portion of samples contain either fewer or more atomic actions.
Longer chunks encode larger egocentric viewpoint shifts and parallax, yielding strong, informative supervision signals for inverse dynamics and supporting long-horizon planning by offering the model an extended temporal context. In contrast, shorter chunks emphasize subtle, fine-grained visual changes (e.g., small turns or local re-alignments), which facilitate precise control and reduce action ambiguity at short horizons. This natural mixture of long and short chunks serves as a complementary curriculum: HPAC contributes global structural cues for planning alongside local adjustment priors for stabilization, thereby improving the model’s understanding of action-grounded visual dynamics and actions and improving robustness to compounding errors during execution.

\begin{figure}[h]
    \centering
    \includegraphics[width=1.0\linewidth]{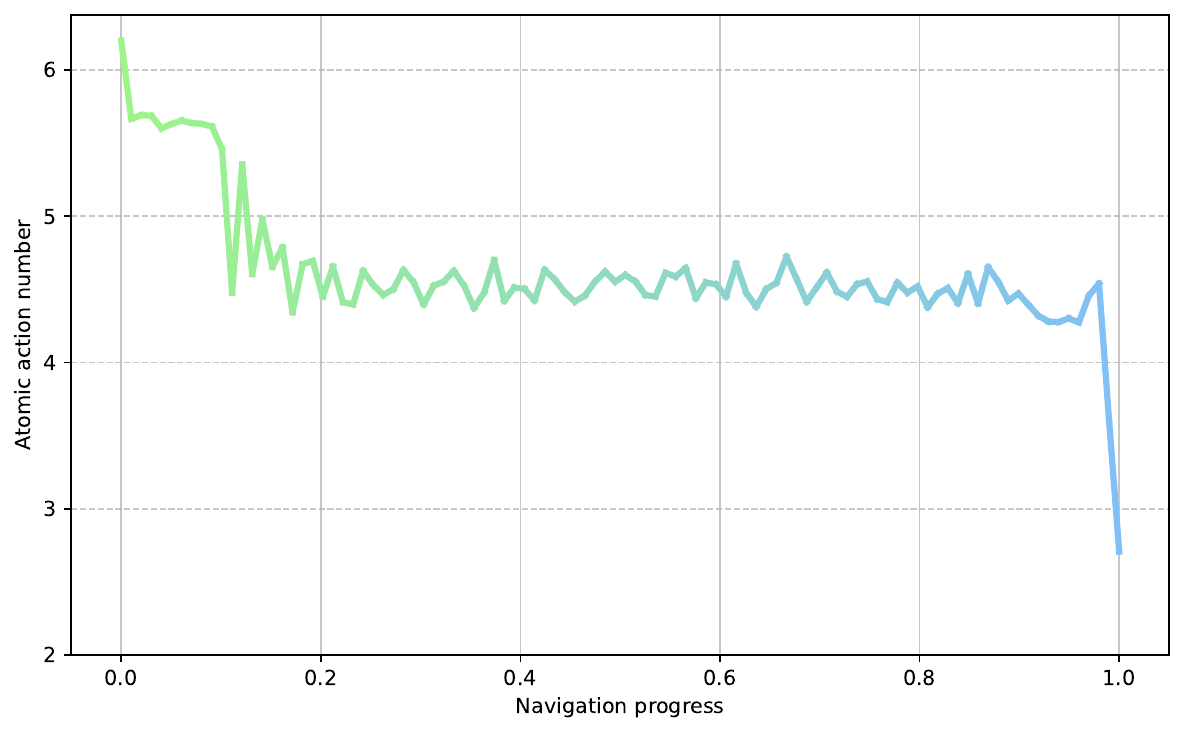}
    \caption{Distribution of actions on navigation progress.}
    \label{fig:data_length_time_step}
\end{figure}

We also normalized the navigation process to a range of 0 to 1, and ~\cref{fig:data_length_time_step} shows the distribution of action chunks as the navigation process progresses. It can be seen that in the early stages of navigation, HPAC merges a large number of atomic actions, which decreases slightly and stabilizes in the middle stages. In the later stages of navigation, closer to the target, HPAC tends to use fewer atomic actions to stabilize the robot's position.

% \begin{figure}[t]
%     \centering
    
%     \begin{subfigure}{0.48\linewidth}
%         \centering
%         \includegraphics[width=\linewidth]{fig/data_distribution.pdf}
%         \caption{Distribution of atomic actions per HPAC chunk in the IDM data.}
%         \label{fig:data_distribution}
%     \end{subfigure}
%     \hfill
%     \begin{subfigure}{0.48\linewidth}
%         \centering
%         \includegraphics[width=\linewidth]{fig/action_length_time_step.pdf}
%         \caption{Distribution of actions on navigation progress.}
%         \label{fig:data_length_time_step}
%     \end{subfigure}
    
%     \caption{Atomic action statistics of HPAC in the dataset.}
%     \label{fig:action_statistics}
% \end{figure}

\subsection{Training Hyperparameters}
\textsc{NaVIDA} is implemented on Qwen-2.5-VL-3B; the vision encoder is frozen and only the projector and LLM are trainable. We use AdamW with lr = 2e-5, BF16, Flash-Attention v2, global batch size = 64; additional hyperparameters are summarized in ~\cref{tab:hyperparams}.

\begin{table}[ht]
\centering
\caption{Training hyperparameters used for fine-tuning Qwen2.5-VL. Only the projector and LLM modules are trainable.}
\label{tab:hyperparams}
\begin{tabular}{l c}
\toprule
\textbf{Hyperparameter} & \textbf{Value} \\
\midrule
Training epochs & $1$ \\
Trainable modules & Projector + LLM  \\
Learning rate & $2e-5$ \\
Optimizer & AdamW \\
AdamW $\beta_1$ & $0.9$  \\
AdamW $\beta_2$ & $0.999$ \\
Adam $\epsilon$ & $1e-8$ \\
Lr schedule &  Linear \\
Global batch size & $64$ \\
Gradient accumulation & $4$ \\
Mixed precision type & BFloat16 \\
Attention type & Flash Attention v2 \\
LLM sequence length & $64800$ \\
Image resolution & $308 \times 252$ \\
\bottomrule
\end{tabular}
\end{table}

\subsection{Real-World Deployment Details}

% ~\cref{fig:real_world_setup} illustrates our deployment system during real-world experiments. Specifically, we deployed  \textsc{NaVIDA}  on a remote server equipped with an RTX 4090. The server and the robot exchanged information via the internet, while the user sent voice commands to the robot via a local area network. The robot acquired RGB data captured by its front-facing camera (shown in ~\cref{fig:robot_setup}), packaged it along with the voice commands, and sent it to the remote server. The server returned motion commands, which were then translated into speed and duration by the robot's motion API to execute the action.

\begin{figure}[h]
    \centering
    \includegraphics[width=0.73\linewidth]{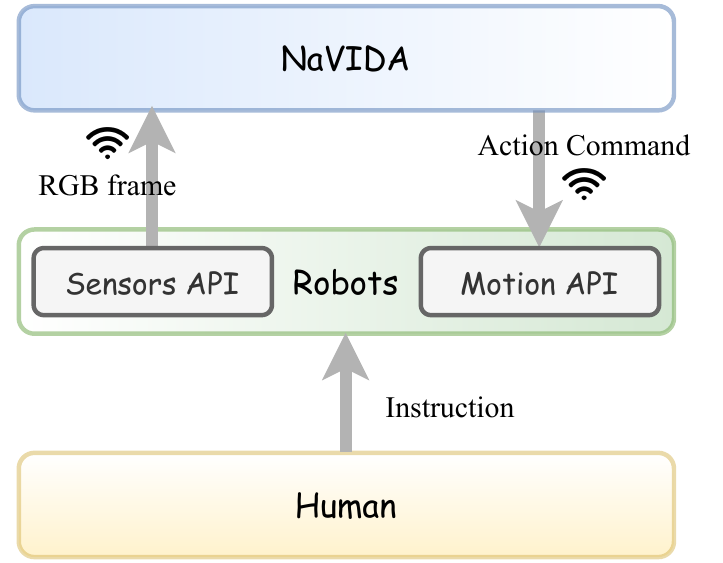}
    \caption{Architecture of the deployed system used for real-world evaluation. RGB frames and user instructions are transmitted from the robot to a remote server running \textsc{NaVIDA}, which predicts action sequences and returns motion commands for execution through the robot’s motion API.}
    \label{fig:real_world_setup}
\end{figure}

~\cref{fig:real_world_setup} shows the overall architecture of our real-world deployment system. Specifically, \textsc{NaVIDA} was deployed on a remote computation server equipped with an NVIDIA RTX 4090 GPU. The system follows a client-server structure, where the robot functions as the client that collects sensory data and executes movement commands, while the remote server is responsible for high-level reasoning and action prediction.

\begin{figure}[h]
    \centering
    \includegraphics[width=0.73\linewidth]{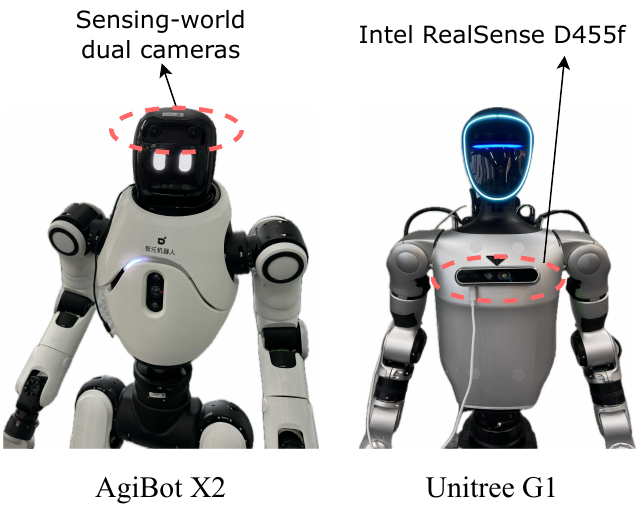}
    \caption{Robot platforms and sensing configuration. AgiBot X2 (left) equipped with a single rectified fisheye RGB camera and Unitree G1 (right) with an additional Intel RealSense D455f front-facing camera, providing consistent monocular RGB input for \textsc{NaVIDA}’s navigation policy.}
    \label{fig:robot_setup}
\end{figure}

During operation, the robot continuously captures RGB images from its front-facing camera (as shown in ~\cref{fig:robot_setup}) and receives natural language or voice commands from the user through a local area network. The sensory data and corresponding commands are then packaged and transmitted to the remote server via the internet. On the server side, the \textsc{NaVIDA} model processes these inputs, performs vision-language reasoning, and predicts the next navigation action. The predicted motion command is then sent back to the robot, where it is converted into executable parameters such as linear velocity, angular velocity, and movement duration through the robot’s motion control API.

In addition, we provide demonstration videos of real-world deployment, which are submitted together with the supplementary materials. The slight latency observed in the videos mainly results from \textbf{network communication delays} between the robot and the remote server.

% \begin{figure}[h]
%     \centering
%     \begin{subfigure}[t]{0.48\linewidth}
%         \centering
%         \includegraphics[width=\linewidth]{fig/real_world_setup.pdf}
%         \caption{Architecture of the deployed system used for real-world evaluation. RGB frames and user instructions are transmitted from the robot to a remote server running \textsc{NaVIDA}, which predicts action sequences and returns motion commands for execution through the robot’s motion API.}
%         \label{fig:real_world_setup}
%     \end{subfigure}
%     \hfill
%     \begin{subfigure}[t]{0.48\linewidth}
%         \centering
%         \includegraphics[width=\linewidth]{fig/robot.pdf}
%         \caption{Robot platforms and sensing configuration. AgiBot X2 (left) equipped with a single rectified fisheye RGB camera and Unitree G1 (right) with an additional Intel RealSense D455f front-facing camera, providing consistent monocular RGB input for \textsc{NaVIDA}’s navigation policy.}
%         \label{fig:robot_setup}
%     \end{subfigure}
    
%     \caption{Real-world evaluation setup.}
%     \label{fig:real_world}
% \end{figure}

\begin{table}[h]
\caption{Ablation study of the vision encoder fine-Tuning on VLN-CE R2R Val-Unseen split.}
\vspace{-2mm}
    \centering
    \begin{tabular}{c|cccc}
    \toprule
    Vision Encoder &  NE$\downarrow$ & OS$\uparrow$ & SR$\uparrow$ & SPL$\uparrow$ \\
    \midrule
      Trainable  & 5.12  & \textbf{63.6} & \textbf{55.6}	& 49.2 \\
      Frozen     & \textbf{4.91}  & 63.2 & 55.5 & \textbf{49.6} \\
      \bottomrule
    \end{tabular}
    \label{tab:ablation_on_vision_encoder}
\end{table}

\section{More Experiment Results}
\label{sec:app_results}

\subsection{More Ablations}
\label{sec:app_ablations}

\begin{figure}[h]
    \centering
    \includegraphics[width=0.75\linewidth]{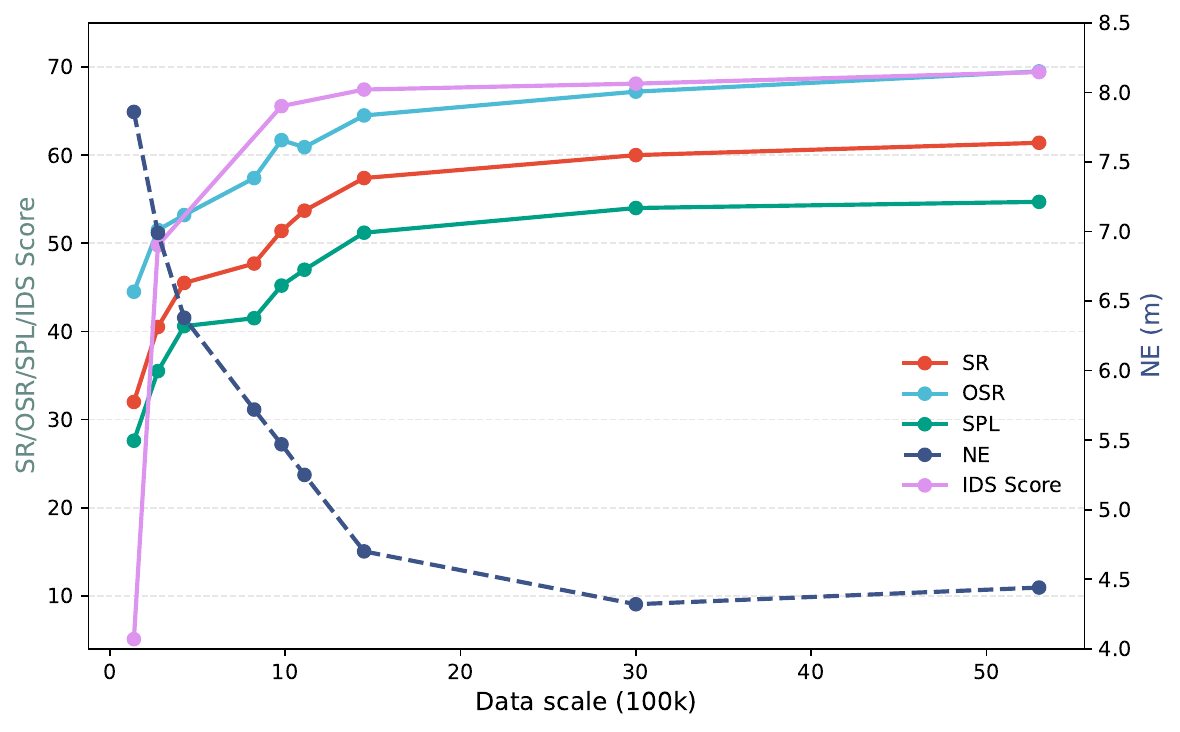}
    \caption{Navigation performance of \textsc{NaVIDA} on VLN-CE R2R Val-Unseen split with the increase of data scale. SR/OSR/SPL/IDS Score increase and NE decreases as data grows, with gains tapering at larger scales.}
    \label{fig:performance-data_scale}
\end{figure}

\noindent \textbf{Data Scale.} ~\cref{fig:performance-data_scale} illustrates the performance trend of \textsc{NaVIDA} on R2R-CE Val-Unseen as a function of the total data size, which includes both IDS and VLN data. We also constructed a validation set for an IDS on R2R-CE Val-Unseen and plotted the variation curve of the IDS score. As the total data size increases, \textsc{NaVIDA} consistently achieves better performance. The improvement is particularly pronounced when the data scale is small, indicating that \textsc{NaVIDA} effectively leverages additional data to learn more robust representations in low-data regimes. However, as the total data scale continues to grow, the performance gains gradually saturate. This behavior likely reflects a diminishing returns effect, where the model has already captured most of the useful patterns from the available data, and further increases in data size contribute less to performance. Additionally, the stabilization may indicate that other factors such as model capacity, task complexity, or inherent noise in the data become the limiting factors once a certain data threshold is reached.

\begin{figure*}[h]
    \centering
    \includegraphics[width=\linewidth]{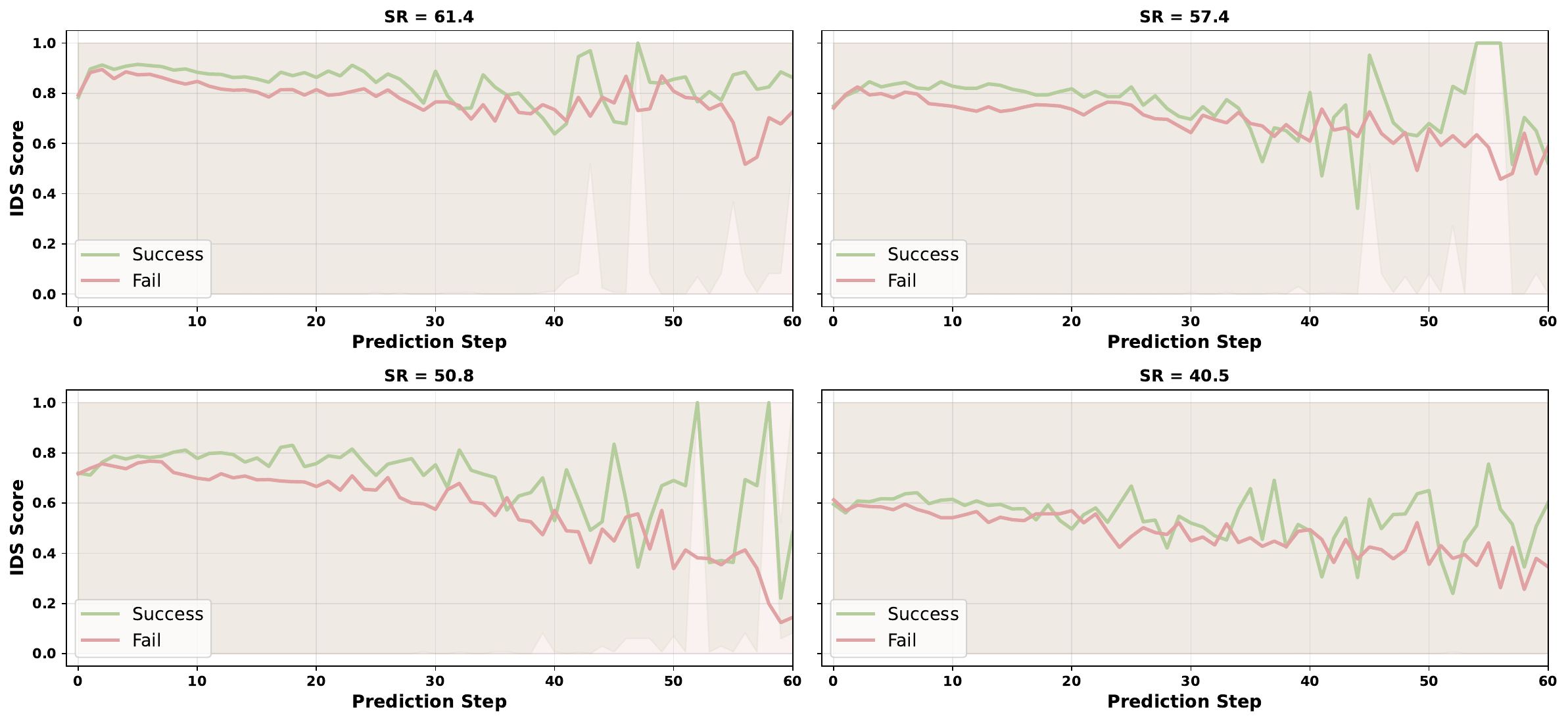}
    \caption{IDS score over prediction steps for successful and failed trajectories. Successful episodes exhibit consistently higher and more stable IDS scores, while failed episodes show gradual degradation, indicating the importance of accurate action-state alignment for long-horizon navigation.}
    \label{fig:ids_score}
\end{figure*}

\noindent \textbf{Correlation Between IDS and Navigation Performance.}
To quantify the consistency between the IDS-inferred and ground-truth action sequences, we define an \textbf{IDS score} based on the discrepancy of their resulting simulator states.
Rolling out the predicted sequence $\hat{\mathbf{a}}$ and the ground-truth sequence $\mathbf{a}$ yields final poses $(\hat{\mathbf{p}}, \hat{\theta})$ and $(\mathbf{p}, \theta)$, where $\mathbf{p} \in \mathbb{R}^3$ denotes position and $\theta$ the yaw angle.
The positional and angular discrepancy is computed as:
\begin{equation}
\Delta_p = \|\hat{\mathbf{p}} - \mathbf{p}\|_2, \quad \Delta_\theta = 
\min \left( 
\left| \hat{\theta} - \theta \right|,\,
2\pi - \left| \hat{\theta} - \theta \right|
\right).
\end{equation}
The IDS score is then defined as:
\begin{equation}
\mathrm{IDS_{Score}} = 
\exp\left(-\frac{\Delta_p}{\sigma_p}\right)
\cdot
\exp\left(-\frac{\Delta_\theta}{\sigma_\theta}\right),
\end{equation}
where $\sigma_p$ and $\sigma_\theta$ control the sensitivity to positional and angular deviations, respectively. 
In our implementation, we set $\sigma_p = 0.1$ and $\sigma_\theta = \pi/12$.

~\cref{fig:ids_score} illustrates the evolution of IDS scores over prediction steps for successful and failed trajectories under different navigation success rates. Several consistent patterns emerge.

At early stages, successful and failed episodes exhibit comparable IDS scores, suggesting that short-horizon action prediction is generally reliable regardless of the final outcome. As trajectories become longer, however, the two curves gradually separate: successful episodes tend to maintain higher IDS scores, whereas failed ones show a progressive decline. This divergence indicates that preserving action-grounded visual consistency becomes increasingly challenging over long horizons.

We further observe that models with higher navigation success rates consistently achieve higher and more stable IDS scores throughout execution. Since IDS scores are computed solely from the agreement between inferred and ground-truth action-induced state transitions, independent of trajectory success labels, this correlation suggests that the metric captures an ability closely associated with effective navigation. In particular, stronger navigation agents tend to maintain more consistent action-conditioned visual dynamics over extended decision horizons, while weaker agents accumulate larger transition inconsistencies as execution proceeds.

\noindent \textbf{IDS Ablation under Matched Data Budgets.}
Since adding IDS increases the total number of training samples, we conduct a controlled ablation to disentangle the effect of IDS from the effect of data scale as shown in Tab \ref{tab: controlled data ablations}. For the non-IDS baseline, we repeat the same data construction process using the same trajectory sources until its total data volume matches that of the IDS setting. We evaluate two matched budgets, 270K and 1.45M samples. Under both settings, IDS consistently improves performance, indicating that the gain comes from action-grounded dynamics supervision rather than simply from seeing more data.

\newcommand{\noidm}{$\circ$}      % 不带 IDM
\newcommand{\withidm}{$\bullet$}  % 带 IDM
\begin{table}[h]
\centering
\begin{tabular}{cc|cccc}
\toprule
Data & Data Size & NE$\downarrow$ & OS$\uparrow$ & SR$\uparrow$ & SPL$\uparrow$ \\
\midrule
\noidm & 270K & 7.34 & 45.1 & 36.5 & 32.4 \\
\withidm & 270K & 6.99 & 51.5 & 40.5 & 35.5 \\
\noidm & 1450K & 4.97 & 62.4 & 53.8 & 47.9 \\
\withidm & 1450K & 4.91 & 63.2 & 55.5 & 49.6 \\
\bottomrule
\end{tabular}
\vspace{-2.5mm}
\caption{Ablation study of controlled data budgets with \withidm \, contains IDS.}
\label{tab: controlled data ablations}
\end{table}

\noindent \textbf{Merge Probabilities.}
We additionally analyze HPAC with merge probabilities of $0.3$, $0.7$, and $1.0$ as is shwon in Tab \ref{tab: merge prob}. The results show that $p=0.7$ achieves the best performance while both lower and higher merge probabilities lead to consistent performance degradation.

\begin{table}[h]
\centering
\begin{tabular}{c|cccc}
\toprule
Merge Prob & NE$\downarrow$ & OS$\uparrow$ & SR$\uparrow$ & SPL$\uparrow$ \\
\midrule
0.3 & 5.27 & 60.3 & 51.1 & 45.9 \\
0.7 & 4.91 & 63.2 & 55.5 & 49.6 \\
1.0 & 5.63 & 58.7 & 49.2 & 42.8 \\
\bottomrule
\end{tabular}
\vspace{-2.5mm}
\caption{Ablation study of merging probabilities.}
\label{tab: merge prob}
\end{table}

\noindent \textbf{Number of History Frames.}
We perform an ablation study to investigate how the number of historical frames influences navigation performance on the R2R-CE Val-Unseen split. All models are trained solely on the R2R training data for consistency. As shown in~\cref{tab:ablation_on_history_frames}, using $8$ history frames is sufficient to capture most instruction horizons, providing the model with adequate temporal context for action reasoning. Increasing the number of frames to $16$ does not yield noticeable performance gains and occasionally leads to a slight degradation, likely due to redundant or noisy visual information accumulated over longer sequences. This behavior aligns with findings from NaVILA~\cite{navila}, indicating that excessive historical context can dilute the relevance of recent observations and hinder temporal grounding. Hence, we adopt $8$ history frames as a balanced choice between contextual richness and model efficiency.

\noindent \textbf{Model Scaling.}
We train NaVIDA with a larger backbone on R2R and DAgger data due to limited time. As shown in Table~\ref{tab: model_size}, a stronger base model consistently improves performance with IDS, indicating that NaVIDA scales well with stronger base models.

\begin{table}[h]
\centering
\begin{tabular}{cc|cccc}
\toprule
\# Model Param & Data & NE$\downarrow$ & OS$\uparrow$ & SR$\uparrow$ & SPL$\uparrow$ \\
\midrule
3B & \noidm & 5.98 & 57.3 & 49.2 & 44.4 \\
3B & \withidm & 5.29 & 57.9 & 50.4 & 44.7 \\
7B & \noidm & 5.65 & 54.3 & 51.3 & 46.8 \\
7B & \withidm & 5.34 & 55.6 & 52.7 & 48.4 \\
\bottomrule
\end{tabular}
\vspace{-2.5mm}
\caption{Ablation study of the model parameters with \withidm \, contains IDS.}
\label{tab: model_size}
\end{table}

\noindent \textbf{Vision Encoder: Frozen vs. Trainable.}
We conduct an ablation study to evaluate the impact of fine-tuning the vision encoder on overall navigation performance. As shown in ~\cref{tab:ablation_on_vision_encoder}, unfreezing the vision encoder and training it end-to-end provides only marginal improvements compared to keeping it frozen. This result suggests that the pre-trained visual representations are already sufficiently informative for downstream navigation tasks, and further adaptation offers limited benefit. Moreover, fine-tuning substantially increases GPU memory usage and training time, with minimal returns in accuracy. Therefore, we keep the vision encoder frozen in all subsequent experiments, enabling the model to efficiently utilize robust pre-trained visual features while maintaining a favorable balance between performance and computational efficiency.

\begin{table}[h]
\caption{Ablation study of the number of history frames on VLN-CE R2R Val-Unseen split.}
\vspace{-2mm}
    \centering
    \begin{tabular}{c|cccc}
    \toprule
    \# History Frames  &  NE$\downarrow$ & OS$\uparrow$ & SR$\uparrow$ & SPL$\uparrow$ \\
    \midrule
      8      & \textbf{6.99} & \textbf{51.5} & \textbf{40.5} & 35.5 \\
      16     & 7.02 & 50.1 & 40.4 & \textbf{36.1} \\
      \bottomrule
    \end{tabular}
    \label{tab:ablation_on_history_frames}
\end{table}

\begin{figure*}[th]
    \centering
    \includegraphics[width=1.0\linewidth]{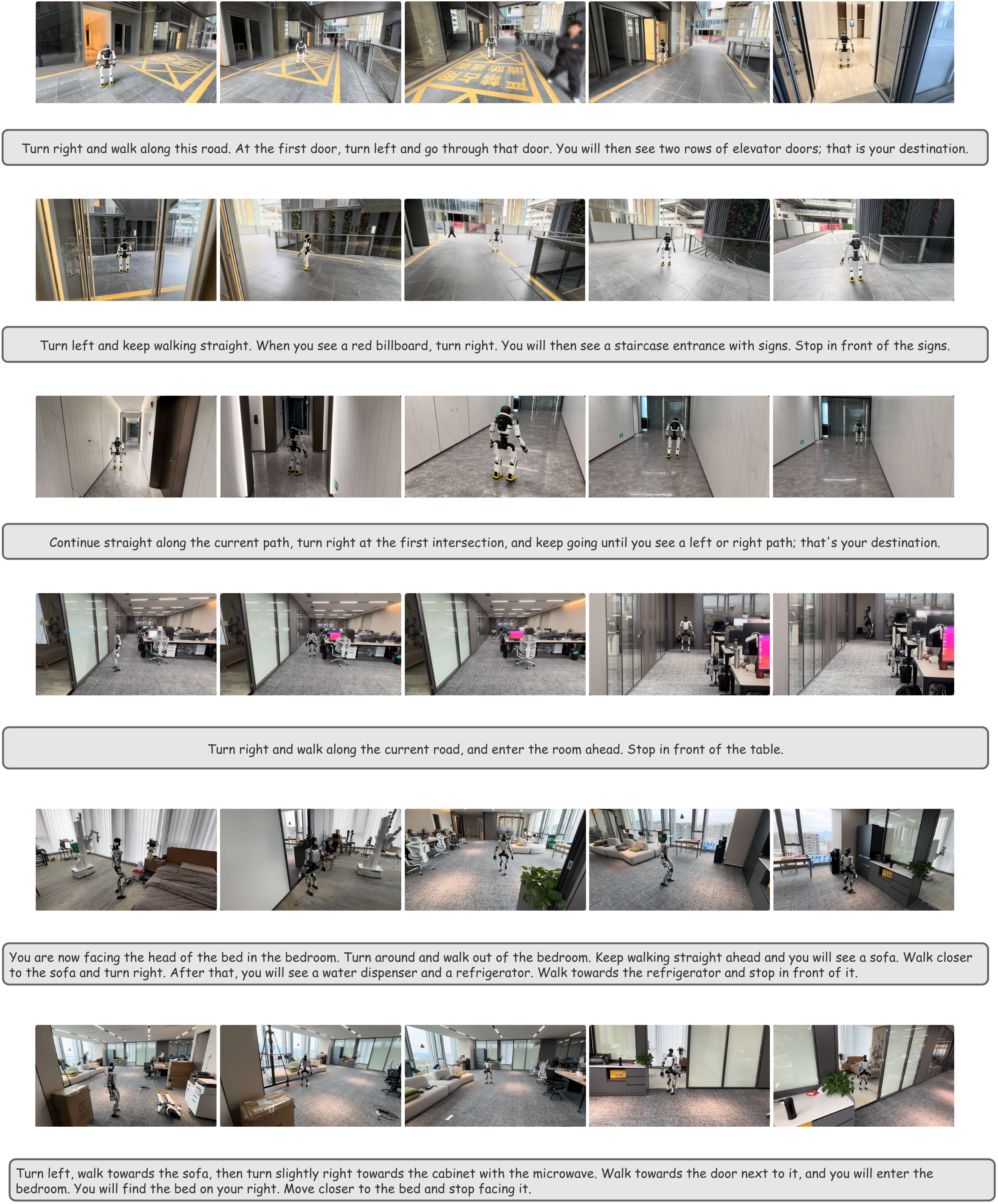}
    \caption{Qualitative results of \textsc{NaVIDA} on real world.}
    \label{fig:appendix_realworld}
\end{figure*}

\begin{figure*}[th]
    \centering
    \includegraphics[width=1.0\linewidth]{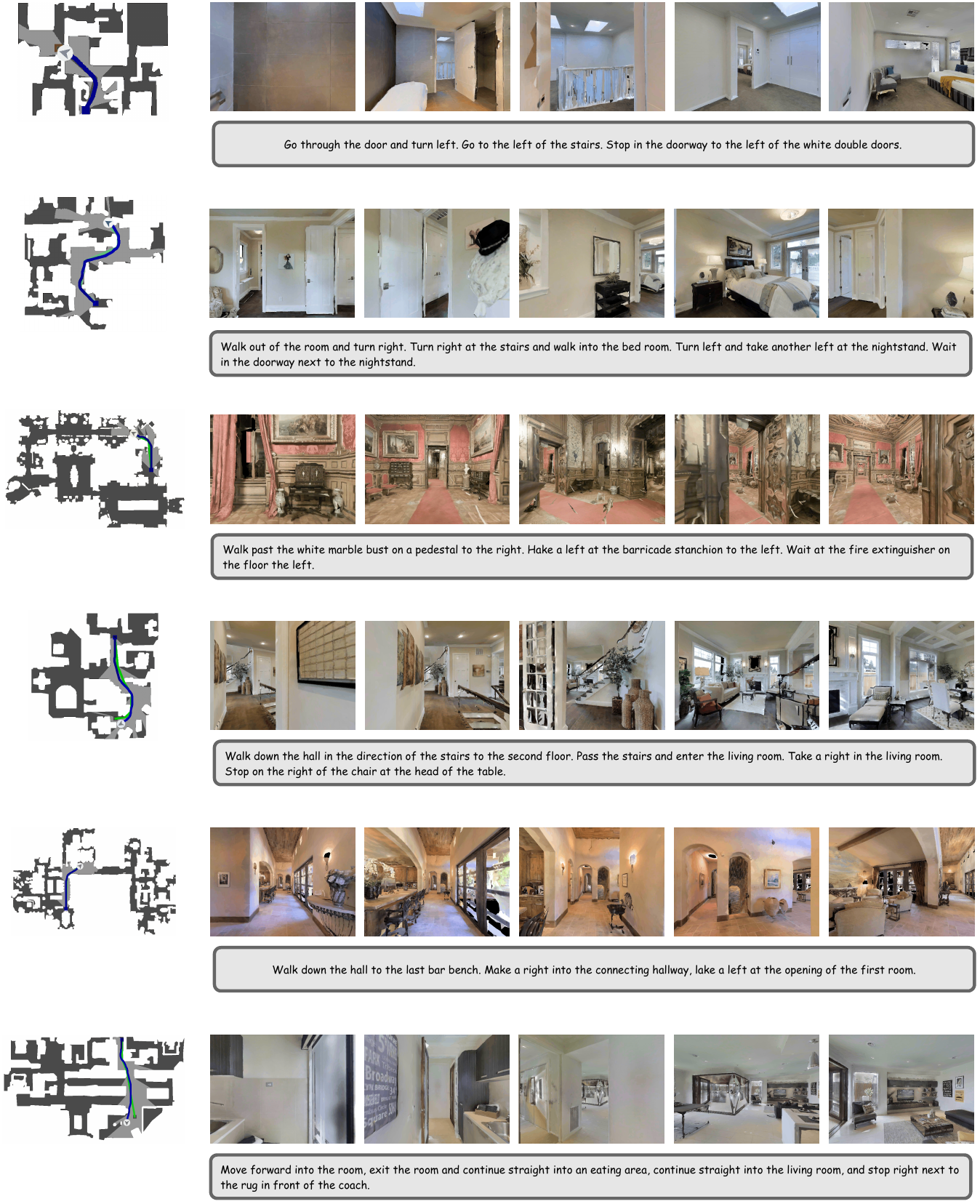}
    \caption{Qualitative results of \textsc{NaVIDA} on VLN-CE R2R Val-Unseen split.}
    \label{fig:appendix_r2r}
\end{figure*}

\begin{figure*}[th]
    \centering
    \includegraphics[width=\linewidth]{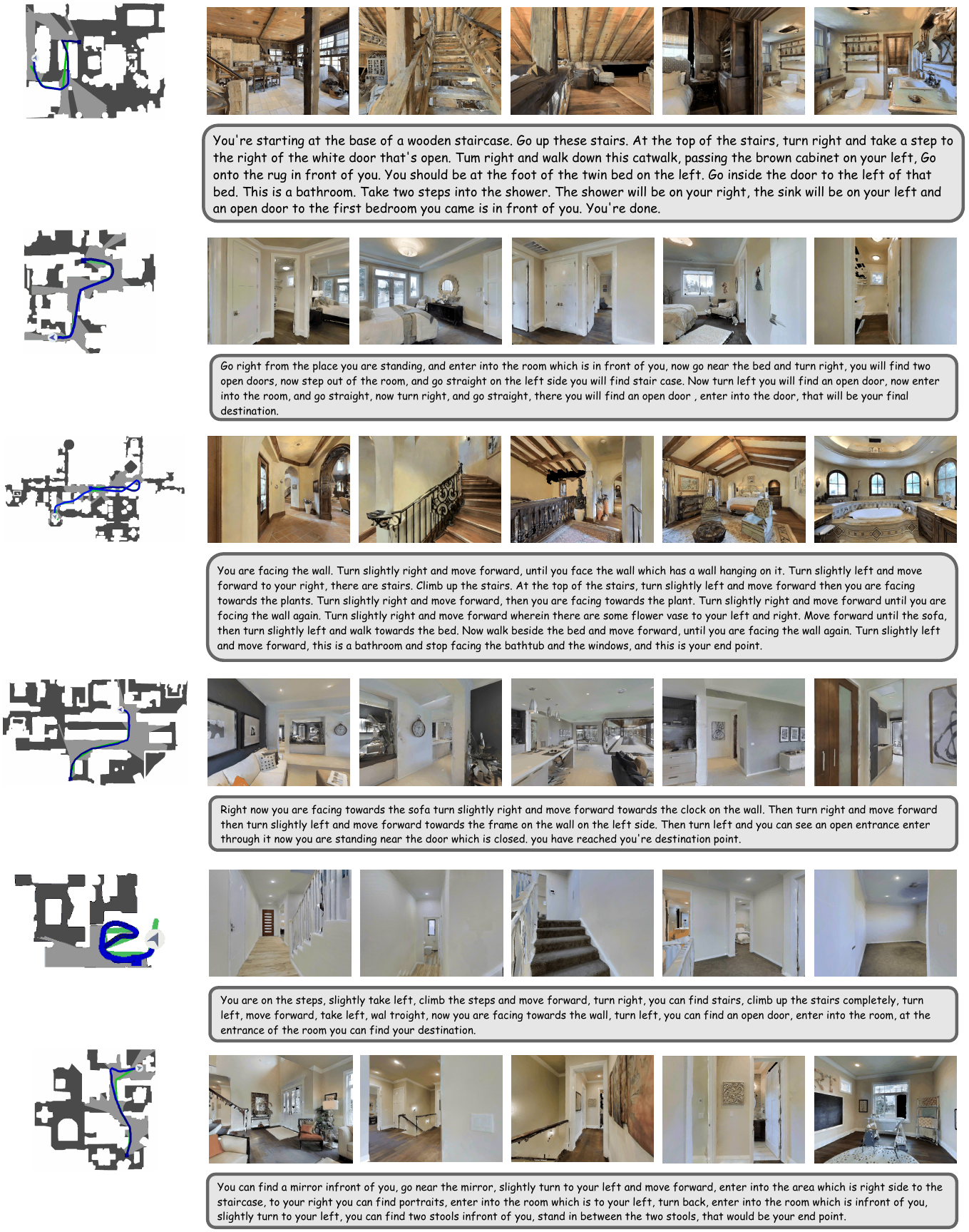}
    \caption{Qualitative results of \textsc{NaVIDA} on VLN-CE RxR Val-Unseen split.}
    \label{fig:appendix_rxr}
\end{figure*}

\subsection{More Qualitative Results}
\label{sec:app_qualitative}

% We present extensive qualitative results of \textsc{NaVIDA} in both real-world and simulated environments to further demonstrate its effectiveness and generalization ability. The results are shown in~\cref{fig:appendix_realworld},~\cref{fig:appendix_r2r} and~\cref{fig:appendix_rxr}
We present extensive qualitative results of \textsc{NaVIDA} in both real-world and simulated environments to further demonstrate its effectiveness and generalization ability, as shown in~\cref{fig:appendix_realworld},~\cref{fig:appendix_r2r}, and~\cref{fig:appendix_rxr}. The results show that \textsc{NaVIDA} produces smooth, instruction-aligned trajectories in simulation and robustly executes navigation commands in real-world settings. These findings highlight the model’s strong cross-domain generalization, accurate language grounding, and reliable decision-making in complex, long-horizon navigation tasks.

% \clearpage  % TODO FINAL: This \clearpage needs to be removed from both review and camera-ready versions.

% % \section*{Acknowledgements}
% % Please insert your acknowledgments here.

% % ---- Bibliography ----
% %
% % BibTeX users should specify bibliography style 'splncs04'.
% % References will then be sorted and formatted in the correct style.
% \bibliography{main}
% \end{document}

\end{document}